\documentclass[acmtog , nonacm=true]{acmart}

\usepackage{booktabs} 
\usepackage{enumitem}

\usepackage{tabularray}
\usepackage{makecell} 
\usepackage[ruled]{algorithm2e} 

\SetAlFnt{\small}
\SetAlCapFnt{\small}
\SetAlCapNameFnt{\small}
\SetAlCapHSkip{0pt}

\setcopyright{rightsretained}
\acmJournal{TOG}
\acmYear{2023} 
\acmVolume{42} 
\acmNumber{6} 
\acmArticle{264} 
\acmMonth{12} 
\acmPrice{15.00} 
\acmDOI{10.1145/3618329} 

\newcommand{\btheta}{\boldsymbol{\theta}}
\newcommand{\bbeta}{\boldsymbol{\beta}}
\newcommand{\btau}{\boldsymbol{\tau}}

\newcommand{\bPhi}{\boldsymbol{\Phi}}
\newcommand{\bPsi}{\boldsymbol{\Psi}}

\newcommand{\br}{\mathbf{r}}

\newcommand{\bI}{\mathbf{I}}

\newcommand{\bV}{\mathbf{V}}

\newcommand{\bp}{\mathbf{p}}

\newcommand{\bc}{\mathbf{c}}

\newcommand{\bJ}{\mathbf{J}}
\newcommand{\ie}{\textit{i.e. }}
\newcommand{\eg}{\textit{e.g. }}
\newcommand{\etal}{\textit{et al. }}
\newcommand{\method}{\textit{Decaf}}
 
\definecolor{darkorange}{rgb}{1.0, 0.55, 0.0}

\begin{document}
\title{Decaf: Monocular Deformation Capture for Face and Hand Interactions} 
\author{Soshi Shimada}
\orcid{0000-0001-8117-5125} 
\affiliation{
\institution{MPI for Informatics, SIC, VIA Research Center}
\city{Saarbrücken} 
 \country{Germany}
}
\email{sshimada@mpi-inf.mpg.de}

\author{Vladislav Golyanik}
\orcid{0003-1630-2006} 
\affiliation{
\institution{MPI for Informatics, SIC}
\city{Saarbrücken} 
 \country{Germany}
}

\email{golyanik@mpi-inf.mpg.de}

\author{Patrick Pérez}
\orcid{0000-0002-8124-1206}
\affiliation{
\institution{Valeo.ai} 
\city{Paris} 
 \country{France}}
\email{patrick.perez@valeo.com}

\author{Christian Theobalt}
\orcid{0000-0001-6104-6625}
\affiliation{
\institution{MPI for Informatics, SIC, VIA Research Center}
\city{Saarbrücken} 
 \country{Germany}
}
\email{theobalt@mpi-inf.mpg.de}
   
\begin{abstract} 
Existing methods for 3D tracking from monocular RGB videos predominantly consider articulated and rigid objects (\textit{e.g.,} two hands or humans interacting with rigid environments). 
Modelling dense non-rigid object deformations in this setting (\eg when hands are interacting with a face), remained largely unaddressed so far, although such effects can improve the realism of the downstream applications such as AR/VR, 3D virtual avatar communications, and character animations. This is due to the severe ill-posedness of the monocular view setting and the associated challenges (\textit{e.g.,} in acquiring a dataset for training and evaluation or obtaining the reasonable non-uniform stiffness of the deformable object). 
While it is possible to na\"{i}vely track multiple non-rigid objects independently using 3D templates or parametric 3D models, such an approach would suffer from multiple artefacts in the resulting 3D estimates such as depth ambiguity, unnatural intra-object collisions and missing or implausible deformations.  

Hence, this paper introduces the first method that addresses the fundamental challenges depicted above and that allows tracking human hands interacting with human faces in 3D from single monocular RGB videos. 
We model hands as articulated objects inducing non-rigid face deformations during an active interaction. 
Our method relies on a new hand-face motion and interaction capture dataset with realistic face deformations acquired with a markerless multi-view camera system. 
As a pivotal step in its creation, we process the reconstructed raw 3D shapes with position-based dynamics and an approach for non-uniform stiffness estimation of the head tissues, 
which results in plausible annotations of the surface deformations, hand-face contact regions and head-hand positions. 
At the core of our neural approach are a variational auto-encoder supplying the hand-face depth prior and modules that guide the 3D tracking by estimating the contacts and the deformations. 
Our final 3D hand and face reconstructions are realistic and more plausible compared to several baselines applicable in our setting, both quantitatively and qualitatively. \textcolor{magenta}{\url{https://vcai.mpi-inf.mpg.de/projects/Decaf}}
\end{abstract}

\begin{teaserfigure}
\centering  
\includegraphics[width=\linewidth]{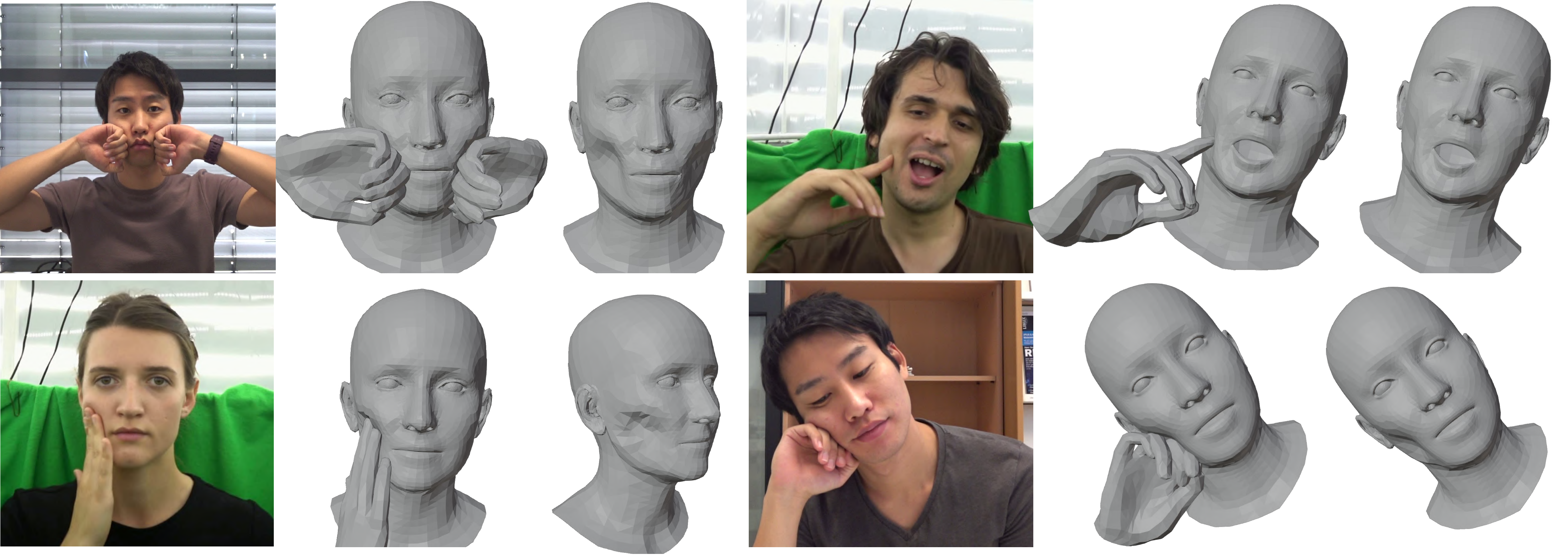} 
\caption{\textbf{Our \method\ approach captures hands and face motions as well as the \textit{face surface deformations} arising from the interactions from a single-view RGB video.} Thanks to our new dataset with 3D surface deformations relying on position based dynamics that 
considers the underlying human skull structure, our neural architecture estimates plausible hands-head interactions and head deformations. The examples in this figure highlight the variety of the supported hand poses and facial expressions. The results are temporally consistent. See our supplementary video for dynamic visualisations. 
\label{fig:teaser} 
} 
\end{teaserfigure} 
%
%


%
%

\maketitle 
\section{Introduction} 
Reconstructing 3D hands and face from a \textbf{monocular RGB video} is a challenging and important research area in computer graphics. The task becomes significantly more difficult when attempting to reconstruct hands and face simultaneously including \textit{surface deformations caused by their interactions}. 
Capturing such interactions and deformations is crucial for enhancing realism in reconstructions as they are frequently observed in everyday life (hand-face interaction occurs $23$ times per hour on average during awake-time \cite{kwok2015face}), and they significantly impact the impressions formed by others. Consequently, reconstructing hand-face interactions is key for avatar communication, virtual/augmented reality, and character animation, where realistic facial movements are essential to create an immersive experience, as well as for applications such as sign language transcriptions and driver drowsiness monitoring. 
Despite several studies on the reconstruction of face and hand motions, the capture of interactions between them and the corresponding deformations from a monocular RGB video remains unaddressed \cite{TretschkNonRigidSurvey}. On the other hand, na\"ively using existing template-based hand and face reconstruction methods leads to artefacts such as collisions, and missing interactions and deformations due to the inherent depth ambiguity in the monocular setting and the lack of deformation modelling in the reconstruction pipeline. 

Several key challenges are associated with this problem setting. 
One (I) is the lack of an available markerless RGB capture dataset for face and hand interaction with non-rigid deformations
for model 
training and method evaluation. Capturing such a dataset is highly challenging due to the constant presence of occlusions caused by hand and head motions, particularly at the interaction region where non-rigid deformation occurs. Another challenge (II) is the inherent depth ambiguity of the single-view RGB setup, which   
makes it difficult to obtain accurate localisation information, resulting in errors that can cause implausible artefacts such as collisions or non-touching of the hand and head (when they interact in practice). To tackle these challenges, we propose \method (short for \textit{deformation capture of faces interacting with hands}), a monocular RGB method for capturing face and hand interactions along with facial deformations. 
Specifically, to address (I), we propose a solution that combines a multiview capture setup with a position-based dynamics simulator for reconstructing the interacting surface geometry, even under occlusions. To integrate the deformable object simulator, we calculate the stiffness values of a head mesh using a simple but effective``skull-skin distance'' (SSD) method. This approach provides non-uniform stiffness to the mesh, which significantly improves the qualitative plausibility of the reconstructed geometry compared with uniform stiffness values. To address the challenge (II), we train the networks to obtain the 3D surface deformations, contact regions on the head and hand surfaces, and the interaction depth prior from single-view RGB images utilising our new dataset. During the final optimisation stage, we utilise this information from different modalities to obtain plausible 3D hand and face interactions with non-rigid surface deformations, which helps disambiguate the depth ambiguity of the single-view setup. Our approach results in much more plausible hands-face interactions compared to the existing works; see Fig.~\ref{fig:teaser} for representative results. 
In summary, the primary technical contributions of this article are as follows: 
\begin{itemize}[topsep=7pt]
\itemsep0em 
\item \method, %
the first learning-based MoCap approach for 3D hand and face interaction reconstruction with face surface deformations (Sec.\,\ref{sec:method}). 
\item %
A global fitting optimisation guided by the estimated contacts, learned interaction depth prior, and deformation model of the face to enable plausible 3D interactions (Sec.~\ref{ssec:fittin_opt}). 
\item The acquisition of the first markerless RGB-based 3D hand-face interaction dataset with surface deformations with consistent topology based on position-based dynamics (PBD). The reference 3D data for model training and evaluation 
are generated using a simple and effective non-uniform stiffness estimation approach for human head models, namely \textit{skull-skin distance} (\textit{SSD}; Sec.~\ref{sec:dataset}). 
\end{itemize} 
Our \method\ outperforms benchmark and existing related methods both qualitatively and quantitatively, with notable improvements in physical plausibility metrics (Sec.~\ref{ssec:quantitative}). 
For dynamic qualitative comparisons, please refer to our supplementary video. 
We plan to release the acquired dataset and code for research purposes. 

\section{Related Works}  
This section focuses on the 3D reconstruction of hands interacting with objects in the monocular (single-view) capture context. 

\subsection{Hand Reconstruction with Interactions}
There have been diverse works proposed to capture 3D hand motions with interactions. Several works reconstruct 3D hand and rigid object interactions from depth information \cite{hu2022physical,zhang2021single,zhang2019interactionfusion} or RGB camera \cite{cao2021reconstructing,liu2021semi,tekin2019h+,grady2021contactopt}. There are several works that reconstruct hand-hand interactions. Mueller \etal \shortcite{mueller2019real} reconstruct two hands interactions from a single depth camera utilising collision proxies based on Gaussian spheres embedded in the hand model. Some works reconstruct interacting 3D hands from a single RGB image \cite{zhang2021interacting,HandFlow_VMV2022}. However, none of these works considers the non-rigidity while interactions unlike ours.

Similar to our approach, Tsoli \etal \shortcite{tsoli2018joint} reconstruct \textbf{non-rigid} cloth and hand interaction by considering hand/object contact points in the optimisation. However, the method requires \textbf{RGB-D} input unlike ours. Our work assumes no access to depth sensor information and reconstructs interactions with a deformable face. The face exhibits varying stiffness values based on the surface area, owing to the underlying skull structure in a human's head. This is in contrast to cloth interactions, which typically have uniform stiffness values. Furthermore, our face autonomously changes its pose and expression during the sequence, whereas in \cite{tsoli2018joint}, the behaviour of the cloth changes only due to the interacting hand or gravity. These unique characteristics, coupled with the limited input setting, make our problem highly challenging.

\begin{figure*}[t!] 
\includegraphics[width=1\linewidth]{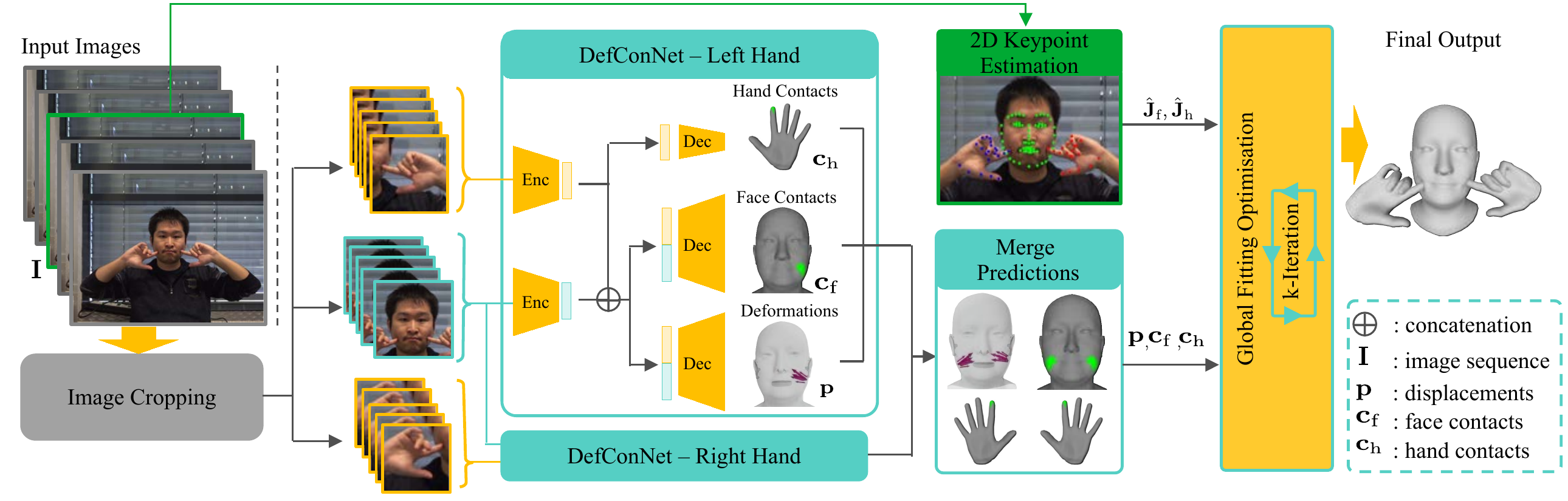} 
\caption{Schematic visualisation of \method, the proposed system to predict 3D poses of hands and face in interaction from a sequence of monocular RGB images of a subject. The input image sequence is first cropped on the left-/right-hand and face locations, which are subsequently fed to the DefConNet, where we estimate the probabilities of face-hand contact $\bc_{\text{f}}$ and $\bc_{\text{h}}$ as well as the per-vertex displacements $\mathbf{p}_{0}$. Two DefConNets of the same architecture are independently operated in cases where two hands are present in the scene. Next, the contact labels and deformations are merged by simply computing the union of the outputs from the two DecConNets. Finally, we solve the optimisation to fit the hands and face parametric models guided by the estimated contacts, deformations and $2$D keypoints in the image (Sec.~\ref{ssec:fittin_opt}). The final output from \method\ reconstructs the face and hands, incorporating plausible surface deformations on the face resulting from their interactions.
}  
\label{fig:schematic_draw}
\end{figure*} 

\subsection{Monocular Face Reconstruction} Capturing a human face from a single view RGB input is important for many graphics applications, thus a significant amount of works have been proposed with learning-free \cite{garrido2016corrective,garrido2013reconstructing,wu2016anatomically,thies2016face2face} and learning-based approaches \cite{ichim2015dynamic,saito2016real, lattas2020avatarme}. In this category, some works train the networks in a self-supervised manner to reconstruct faces with textures and illuminations \cite{tewari17MoFA} or details with estimated normals \cite{EMOCA:CVPR:2021,feng2021learning}. Although these works capture the geometry of expressive deforming human faces, none of the works in this category models the face deformations caused by the interactions unlike ours.

\subsection{Shape from Template (SfT)} 
This algorithm class bears a similarity to our approach. SfT assumes a template mesh of the tracking object and deforms the template mesh based on the observations such as RGB/-D sequences. Several works address this problem with  learning-based algorithms \cite{bozic2020neuraltracking,shimada2019ismo,golyanik2018hdm,FuentesJimenez2021,Kairanda2022}, and some with learning-free optimisation-based approaches \cite{ NRST_GCPR2018, Ngo2015,Salzmann2007,yu2015direct,zollhofer2014real}. Unlike these approaches, our method models \textit{interactions} between two different objects (\ie hand and face) from a single view RGB input under severe occlusions caused by the interactions. Petit \etal \shortcite{petit2018capturing} propose a physics-based non-rigid object tracking method using a finite element method. However, their method requires RGB-D input and focuses on simple deformable objects (\eg, cubes and discs). In contrast, our approach does not rely on depth information and handles interactions between a complex articulated hand and face, considering locally varying stiffness values. Some works estimate 3D human poses with self- and multi-person interactions (contacts) from single RGB images \cite{Mueller:CVPR:2021,fieraru2021learning,fieraru2020three}. However, they do not model significant surface deformations due to 
contacts (\eg\ during hand-face interactions). 
Li \etal \shortcite{Li_3DV2022} propose a method that addresses a problem set that bears resemblance to ours. It estimates the $3$D global human pose along with the deformations of the interacting environment surface based on ARAP-loss. However, their method does not consider stiffness values specific to object categories and does not incorporate learned priors for non-rigid deformations, distinguishing it from our approach.
\subsection{Template Free Non-Rigid Surface Tracking} 
Some methods in this category reconstruct non-rigid surfaces by acquiring first an explicit template mesh from RGB-D inputs \cite{innmann2016volumedeform}. Some use node graphs \cite{lin2022occlusionfusion} or implicit SDF surface representations  \cite{ slavcheva2017killingfusion} for non-rigid surface tracking. Guo \etal \shortcite{guo2017real} propose a method that reconstructs the non-rigid surface along with the surface albedo and low-frequency lighting. Our approach differs from these works by considering the dynamics of the interactions between two different materials \ie face and hand, and face surface stiffness values based on bone structure. Additionally, our dataset and method's output have consistent 3D mesh topologies that are very important for the supervision of network training in explicit surface space.

\subsection{Physics-based MoCap}
Recently, numerous physics-based algorithms for motion capture have been proposed. Several works model the interactions with the environment from a static single RGB camera \cite{PhysAwareTOG2021,PhysCapTOG2020,rempe2020contact,gartner2022trajectory,gartner2022differentiable,xie2021physics,luo2022embodied,huang2022neural,innmann2016volumedeform,yuan2021simpoe} or with objects \cite{GraviCap2021}. Some works reconstruct 3D poses from egocentric views \cite{luo2021dynamics} or IMUs \cite{PIPCVPR2022}. Hu \etal \shortcite{hu2022physical} reconstruct hand-object interactions from an RGB-D camera sequence modelling the physics-based contact status. 
While the existing approaches primarily focus on modelling the interactions with static floor planes or rigid objects, our method uniquely addresses non-rigid deformations arising from interactions between hands and face. This capability is made possible thanks to our networks trained on our novel dataset, which incorporates 3D deformations generated using a maker-less multiview motion capture system combined with position based dynamics (PBD) \cite{pbd2007} -- a widely adopted deformable object simulation algorithm employed in modern physics engines.

\section{Method}
\label{sec:method} 

Our goal is to reconstruct hands interacting with a face in 3D, including non-rigid face deformations caused by the interaction, from a single monocular RGB video. Figure \ref{fig:schematic_draw} provides an overview of the proposed framework. 
Our deformation and contact estimation network \textit{DefConNet}, trained on our new dataset (Sec.\,\ref{sec:dataset}), estimates face surface deformations and contact labels on both face and hand surfaces from an image sequence; the contact labels are crucial to achieve plausible and realistic interactions in 3D 
(Sec.~\ref{ssec:interaction_est}). The estimated deformations, contacts and 2D keypoints are subsequently sent to the global fitting optimisation stage (Sec.~\ref{ssec:fittin_opt}), where we also utilise the \textit{interaction prior} obtained from a conditional variational autoencoder \cite{sohn2015learning} conditioned on the 2D key points for the improved interactions between the hands and face. After this stage, we obtain the final 3D reconstruction of the face and hands in the form of parametric hand and head models with applied deformations. We next explain the notations and assumptions 
used in this work (Sec.~\ref{ssec:modeling}), followed by the details  of our \method approach. 
\begin{figure}[t!] 
\includegraphics[width=1.0\linewidth]{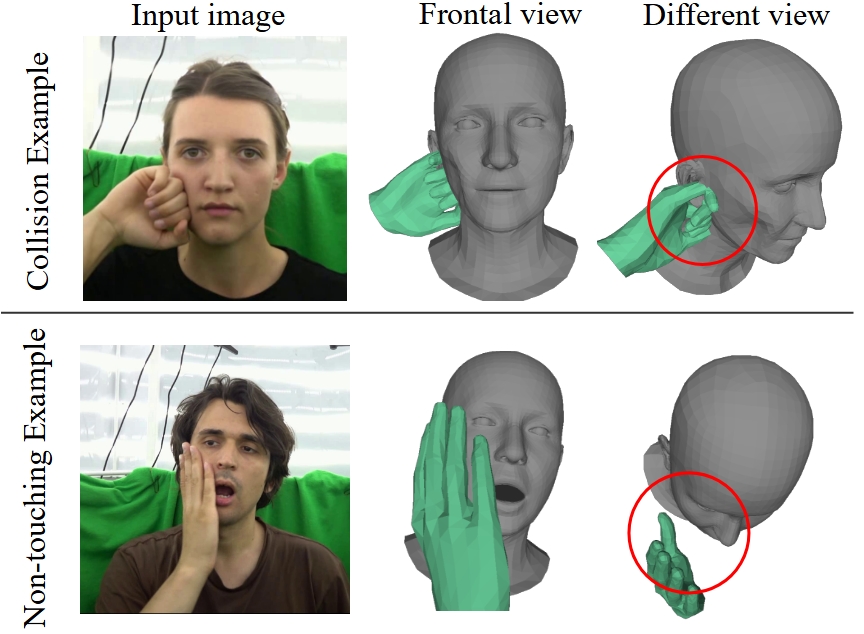} 
\caption{Example artefacts caused by the depth inaccuracies after solving a na\"ive single RGB based fitting optimisation, \ie Eqs. \eqref{eq:fitting_head} and \eqref{eq:fitting_hand} without $\mathcal{L}_{\text{touch}}, \mathcal{L}_{\text{col.}} \,\text{and}\, \mathcal{L}_{\text{depth}}$. The first row shows the physically implausible collisions between the hand and face. The second row displays the ``non-touching'' artefacts in which no hand-face interactions are discernible in the reconstruction, despite the presence of such interactions in the image input. The locations of the artefacts are indicated by the red circles. } \label{fig:artifacts}
\end{figure}

\subsection{Modelling and Preliminaries}\label{ssec:modeling} 
Our \method\ accepts as input a sequence $\bI\,{=}\,\{\bI_t\}\,{=}\,\{\bI_{1},...,\bI_{T}\}$ of $T\,{=}\,5$ successive RGB frames from a static camera with known intrinsic camera parameters. We resize 
$\bI_t$ to $224\,{\times}\,224$ pixels after cropping the detected bounding box around the subject's face and hands in each frame. To represent the 3D face, we employ a gender-neutral version of FLAME parametric model $\mathcal{F}$ \cite{FLAME:SiggraphAsia2017}. We utilise its identity parameters $\bbeta_{\text{f}}\,{\in}\,\mathbb{R}^{100}$, jaw pose $\btheta_{\text{f}}\,{\in}\,\mathbb{R}^{3}$ and expression parameters $\bPsi\,{\in}\,\mathbb{R}^{50}$ combined with the global translation $\btau_{\text{f}}\,{\in}\,\mathbb{R}^{3}$ and rotation $\br_{\text{f}}\,{\in}\,\mathbb{R}^{3}$ that can be formulated as a differentiable function $\mathcal{F}(\btau_{\text{f}}, \br_{\text{f}}, \bbeta_{\text{f}}, \btheta_{\text{f}},\bPsi)$. %
Model $\mathcal{F}$ returns 3D head vertices $\bV_{\text{f}} \,{\in}\,\mathbb{R}^{M\times 3}$ ($M=5023$) from which we obtain the 3D face landmarks $\bJ_{\text{f}} \,{\in}\,\mathbb{R}^{K_{\text{f}}\times 3}$ ($K_{\text{f}}=68$). 
To represent 3D hands, we employ the gender neutral version of the statistical MANO parametric hand model \cite{MANO:SIGGRAPHASIA:2017} that defines the hand mesh as a function $\mathcal{M}(\btau_{\text{h}}, \br_{\text{h}},\btheta_{\text{h}},\bbeta_{\text{h}})$ of global translation $\btau_{\text{h}}\,{\in}\,\mathbb{R}^{3}$ and global root orientation $\br_{\text{h}}\,{\in}\,\mathbb{R}^{3}$, pose parameters $\btheta_{\text{h}}\,{\in}\,\mathbb{R}^{45}$ and hand identity parameters $\bbeta_{\text{h}}\,{\in}\,\mathbb{R}^{10}$. This function $\mathcal{M}$ returns hand 3D mesh vertices $\bV_{\text{h}} \,{\in}\,\mathbb{R}^{N\times 3}$ ($N=778$) from which 3D hand joint positions $\bJ_{\text{h}} \,{\in}\,\mathbb{R}^{K_{\text{h}}\times 3}$ ($K_{\text{h}}=21$) are obtained. We assume that the face identity and hand shape parameters are known. In the following, $\bPhi_{\text{f}} = (\btau_{\text{f}},\br_{\text{f}}, \bbeta_{\text{f}}, \btheta_{\text{f}},\bPsi)$ and $\bPhi_{\text{h}} = (\btau_{\text{h}},\br_{\text{h}}, \bbeta_{\text{h}},\btheta_{\text{h}})$ denote the kinematic states of the face and hand in a 3D space.
 
\begin{figure}[t!] 
\includegraphics[width=1\linewidth]{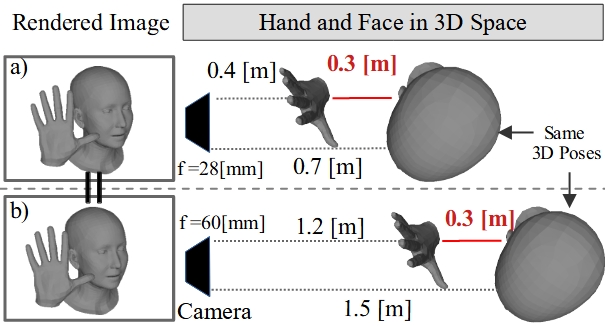} 
\caption{Schematic visualisation of depth ambiguity in a monocular setup. $\text{f}$ denotes the focal length of the camera. 
\textbf{a) and b):} Given the same $3$D poses of face and hand of the same scale in the $3$D space, different combinations of depths and focal lengths can result in indistinguishable images after the $2$D projection in a monocular setting. This effect, known as depth ambiguity, poses a challenge for methods attempting to estimate the depth values of the hand and face in the camera frame from monocular 2D inputs (\textit{e.g.,} RGB images or 2D keypoints). 
However, the relative location of the hand w.r.t. the head is invariant to the positions of the face and hand in $3$D space (\textit{e.g.,} 0.3 [m] above). Based on this idea, our DePriNet learns the depth prior in the \bf{canonical face frame}.}

\label{fig:depth_prior}
\end{figure} 
 \subsection{Interaction Estimation}\label{ssec:interaction_est} 
 We introduce a learning-based approach that estimates plausible interactions in a scene, \textit{i.e.,} the vertex-wise face deformations and 
 contacts on the face and hand surfaces given only single-view RGB images. 
 The approach is trained on our new dataset (Sec.~\ref{sec:dataset}).

 Our neural network accepts as input an image sequence $\bI$ and outputs the deformation on the head model as per-vertex displacements in a camera frame $\bp \,{\in}\,\mathbb{R}^{M\times 3}$, contact labels on the face $\bc_{\text{f}} \,{\in}\,\{0,1\}^M$ and the hand $\bc_{\text{h}} \,{\in}\,  \{0,1\}^N$.
 The contact labels are binary signals \ie~$1$ for contact, $0$ otherwise. The network is trained to estimate the contact probability using the binary cross entropy (BCE):
\begin{equation} 
\label{eq:con_train_loss}
   \mathcal{L}_{\text{labels}} =  
    \operatorname{BCE}(\bc_{\text{f}},\hat{\bc}_{\text{f}})+\operatorname{BCE}(\bc_{\text{h}},\hat{\bc}_{\text{h}}),
\end{equation} 
where $\hat{\bc}_{\text{f}}$ and $\hat{\bc}_{\text{h}}$ denote the ground-truth contact labels for the face and hand, respectively. We also train the network to estimate the deformations using the ground-truth annotations $\hat{\bp}_{m}$: 
\begin{figure*}[t] 
\includegraphics[width=1\linewidth]{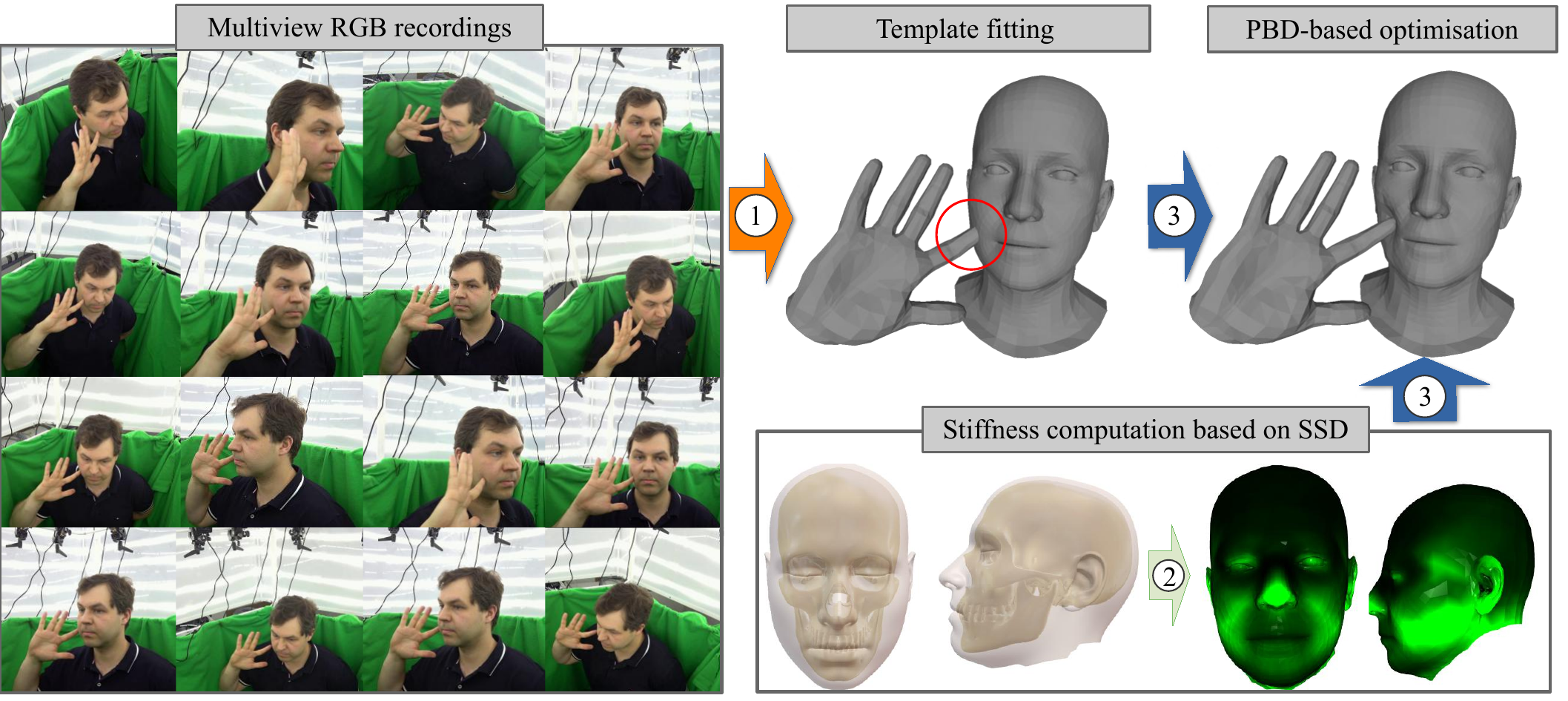} 
\caption{\textbf{Overview of the dataset generation pipeline.} We first capture the hand and face interactions using a markerless multi-view setup. \textcolor{red}{\textbf{(1)}} Subsequently, the obtained RGB image sequences are used to solve template-based fitting optimisation using MANO \cite{MANO:SIGGRAPHASIA:2017} and FLAME \cite{FLAME:SiggraphAsia2017} models. At this stage, due to the unawareness of the hand-face interactions, the collisions are observable, as indicated by the red circle. \textcolor{red}{\textbf{(2)}} To provide the plausible stiffness values on the head mesh for the later position-based dynamics (PBD) optimisation stage, we compute skull-skin distances (SSD) and obtain vertex-wise stiffness values. 
(Left-hand side): A visualisation of mean skull and skin surface of a statistic model \cite{achenbach2018multilinear} from the side and frontal views. (Right-hand side): Transferred stiffness value to FLAME head model \cite{FLAME:SiggraphAsia2017} based on the SSD calculation, see Sec. ~\ref{ssec:stiffness_on_head} for the details. \textcolor{red}{\textbf{(3)}} Using the fitted templates from (1) and the stiffness values from (2), we solve the PBD-based tracking optimisation. This stage handles the physically implausible collisions and provides plausible surface deformations on the head mesh surface (Sec.~\ref{ssec:pbd}).} \label{fig:data_pipeline} 
\end{figure*}
\begin{equation} 
\label{eq:disp_loss}
   \mathcal{L}_{\text{def.}} =\frac{1}{M}\sum^{M}_{m=1} (w^{m}_{\text{def}}\,\big\|\bp_{m}-\hat{\bp}_{m}\big\|_2^{2} 
 + b^{m}_{\text{def}}\,\big\|\bp_{m}\big\|),  
\end{equation} 
where 
\begin{equation} 
  w^{m}_{\text{def}} = 
    \begin{cases} 
     0.3, & \mbox{if} \;\;\; ||\hat{\bp}_{m}|| = 0,\\  
     1.0, & \mbox{otherwise},
    \end{cases} \quad
  b^{m}_{\text{def}} = 
    \begin{cases} 
     1, & \mbox{if} \;\;\; || \bp_{m}|| > \psi,\\  
     0, & \mbox{otherwise}.
    \end{cases} 
\end{equation} 
The first term in Eq. \eqref{eq:disp_loss} allows the network to learn the 3D deformations in our dataset. The weight  $w_{\text{def}}$ helps to penalise the network predictions more on deforming vertices. We observe that this weighting strategy improves the network precision as the majority of the face vertices have no deformations. The second loss term in Eq. \eqref{eq:disp_loss} regularises the unnaturally large deformations on the face surface where $b_{\text{def}}$ works as a binary label to penalise only the vertices with deformations greater than $\psi = 0.1$ [m].

\begin{figure*}[t] 
\includegraphics[width=1\linewidth]{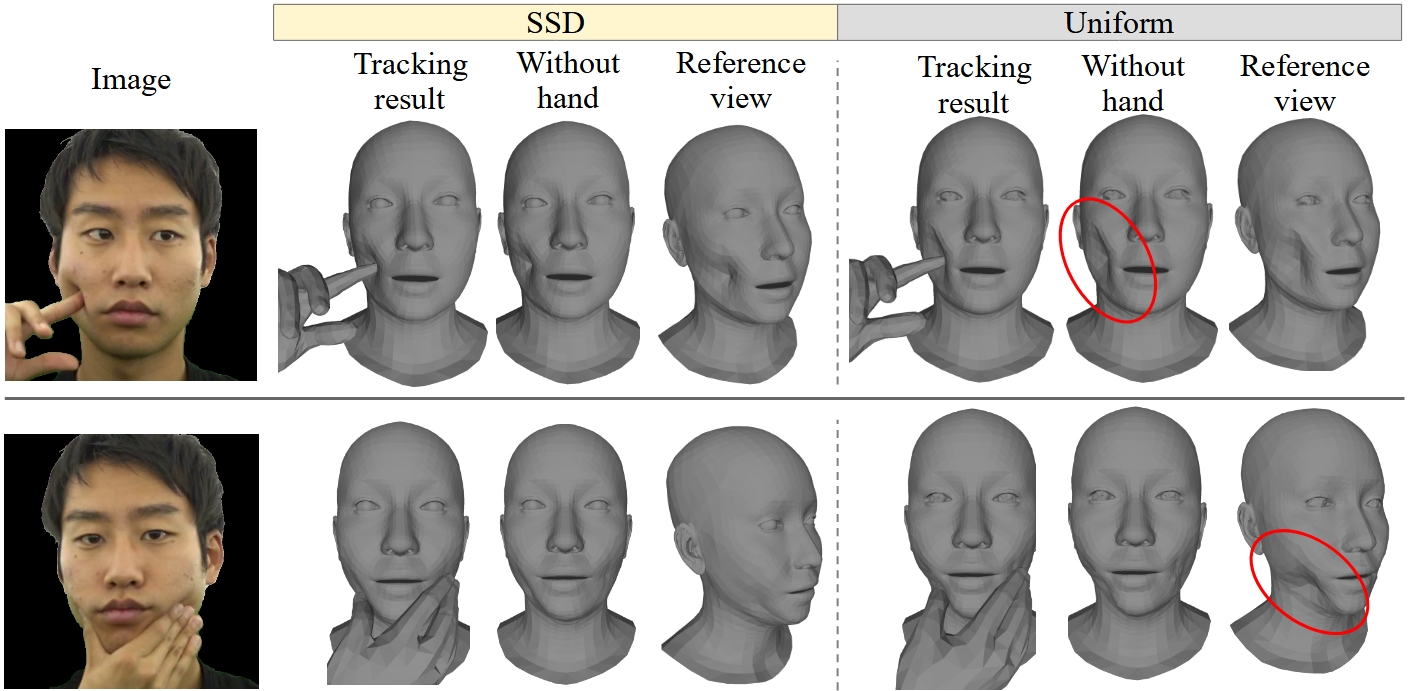} 
\caption{ Example visualisations of the reconstructed 3D head and hand interactions with the stiffness values computed using the skull-skin distance (SSD) (second to fourth columns) and the uniform stiffness value (fifth to seventh columns). With SSD, the obtained surface deformations are much more plausible compared to na\"ievely assigning the uniform stiffness value to all the head vertices.
The red circles highlight the overly deformed surfaces (left) and inaccurate deformations that ignore the underlying jaw in the human head (right). 
} \label{fig:stiffness_combined} 
\end{figure*} 

\subsection{Global Fitting Optimisation}
\label{ssec:fittin_opt} 

Using the estimated deformations $\bp$, contact labels $\bc_{\text{f}}$ and $\bc_{\text{h}}$ and 2D joint keypoints, we obtain the global positions of the face $\bPhi_{\text{f}}$ and hand $\bPhi_{\text{h}}$ in the 3D scene considering their interactions. In this optimisation step, we also update $\bp$ to refine and handle the minor collisions. The objective follows: 
\begin{equation} \label{eq:fitting_opt}
  \mathcal{L}_{\text{opt}}(\bPhi_{\text{f}},\bPhi_{\text{h}}, \bp)  = \mathcal{L}_{\text{face}}+\mathcal{L}_{\text{hand}}.
\end{equation} 
The fitting loss term of the face model $\mathcal{L}_{\text{face}}$ reads: 
\begin{equation} \label{eq:fitting_head}
  \mathcal{L}_{\text{face}}(\bPhi_{\text{f}}, \bp)  =  \mathcal{L}_{\text{2D}} +  \mathcal{L}_{\text{reg.}},
\end{equation} 
where $\mathcal{L}_{\text{2D}}$ and $\mathcal{L}_{\text{reg.}}$ are the weights of the $2$D reprojection term and regulariser loss term  , respectively. Employing the projection function $\Pi(\cdot)$ with the known camera intrinsics, the $2$D reprojection loss term is formulated as follows:
\begin{equation}\label{eq:reproj}
    \mathcal{L}_{\text{2D}} =\frac{1}{M}\sum^{M}_{m=1} w^{m}_{\text{conf.}}\big\|\Pi(\bJ_{\text{f}}^{m}) - \hat{\bJ}_{ \text{f}}^{m}\big\|_2^{2},
\end{equation} where $\hat{\bJ}^{m}_{ \text{f}}$ and $w^{m}_{\text{conf.}}$ are, respectively, the reference $2$D face landmarks and the corresponding confidence value obtained by the method of \cite{bulat2017far} given the input image. We also minimise the regulariser loss term $\mathcal{L}_{\text{reg.}}$ to introduce the statistical prior for the shape $\bbeta_{\text{f}}$ and expression $\bPsi$, and temporal smoothness in the motion:
\begin{equation}\label{eq:reg}
    \mathcal{L}_{\text{reg.}} = \lambda_{\bbeta}\big\|\bbeta_{\text{f}}\big\|_2^{2}+ \lambda_{\bPsi}\big\|\bPsi\big\|_2^{2} + \lambda_{\dot{\bV}}\big\|\dot{\bV}_{\text{f}}  \big\|_2^{2}+\lambda_{\ddot{\bV}}\big\|\ddot{\bV}_{\text{f}} \big\|_2^{2},
\end{equation} where $\dot{\bV}_{\text{f}}$ and $\ddot{\bV}_{\text{f}}$ denote the velocity and acceleration of the head vertex positions $\bV_{\text{f}}$, respectively. $\lambda_{\bullet}$ denotes a weight of the loss term. The objective for the hand fitting $\mathcal{L}_{\text{hand}}$ optimisation includes the $2$D reprojection term $\mathcal{L}_{\text{2D}}$, regulariser term $\mathcal{L}_{\text{reg.}}$, collision term $\mathcal{L}_{\text{col.}}$, \textit{touchness} term $\mathcal{L}_{\text{touch}}$ and the depth prior term $\mathcal{L}_{\text{depth}}$: 
\begin{align}  
\label{eq:fitting_hand}
\begin{split}
  \mathcal{L}_{\text{hand}}(\bPhi_{\text{h}}, \bp)  =   \mathcal{L}_{\text{2D}}+ \mathcal{L}_{\text{reg.}}+  \lambda_{\text{touch}}\mathcal{L}_{\text{touch}}  \\ +\lambda_{\text{col.}} \mathcal{L}_{\text{col.}}+\lambda_{\text{depth}} \mathcal{L}_{\text{depth}}, 
  \end{split}
\end{align} 
where $\lambda_{\bullet}$ are the corresponding weights. The terms $\mathcal{L}_{\text{2D}}$ and $\mathcal{L}_{\text{reg.}}$ are the same as in \eqref{eq:reproj}-\eqref{eq:reg} with the modification that \eqref{eq:reproj} is applied on the hand $3$D joints $\bJ_{\text{h}}$ compared with the reference 2D hand keypoints $\hat{\bJ}_{\text{h}}$, and \eqref{eq:reg} %
on the hand shape $\bbeta_{\text{h}}$, velocity and acceleration of hand vertices, excluding the expression prior loss %
$\big\|\bPsi\big\|_2^{2}$. %
Due to the inaccuracy of the depth estimation in the monocular setting, simply solving the fitting optimisation w.r.t. the face and hand global positions can cause implausible artefacts, \eg collisions between the face and hand or non-touching artefacts. 
Figure \ref{fig:artifacts} shows examples of such artefacts, when solving a na\"ive $2$D reprojection based single view fitting optimisation \ie \eqref{eq:fitting_opt} excluding $\mathcal{L}_{\text{touch}}$, $\mathcal{L}_{\text{col.}}$ and $\mathcal{L}_{\text{depth}}$. They immediately give the impression of unnatural hand-face interaction to the viewer. To address the ``non-touching'' artefacts, we utilise the \textit{touching} loss term $\mathcal{L}_{\text{touch}}$ that penalises the distances between the contact surfaces on the face and hands inspired by \cite{shimada2022hulc}. Specifically, we treat the face and hand vertices with contact probabilities $\bc_{\text{f}}\,{>}\,0.5$ and $\bc_{\text{h}}\,{>}\,0.5$ as effective contacts, respectively. Let $\mathcal{C}_{\text{f}}\,{\subset}\,\llbracket 1,n \rrbracket$ and  $\mathcal{C}_{\text{h}}\,{\subset}\,\llbracket 1,m \rrbracket$ be the index subsets of the face and hand vertices with the effective contacts. 
Using a Chamfer loss,  $\mathcal{L}_{\text{touch}}$ is formulated as follows: 
\begin{equation} \label{eq:touch_term} 
 \mathcal{L}_{\text{touch}} \!=\!\frac{1}{\left|\mathcal{C}_{\text{f}}\right|} \! \sum_{i \in \mathcal{C}_{\text{f}}} \min _{j \in \mathcal{C}_{\text{h}}} \left\|\bV_{\text{f}}^{i}-\bV_{\text{h}}^{j} \right\|_2^2\!\! + \frac{1}{\left|\mathcal{C}_{\text{h}}\right|} \! \sum_{j \in \mathcal{C}_{\text{h}}} \min _{i \in \mathcal{C}_{\text{f}}} \left\|\bV_{\text{f}}^{i}-\bV_{\text{h}}^{j} \right\|_2^2. 
\end{equation} 
To avoid collisions between hands and a head, we also introduce the collision loss term $\mathcal{L}_{\text{col.}}$ for minimising the penetration distance of the hand vertices. 
Specifically, we first detect the hand vertices colliding with the face mesh based on an SDF criterion \cite{sdfgit}.  
Then, we minimise the distance between colliding hand vertices and their nearest vertices on the head mesh. 
Let $\mathcal{P}\,{\subset}\,\llbracket 1,W \rrbracket$ be the subset of indices of hand vertices $\bV_{\text{h}}$ colliding with the face mesh. 
The collision loss is formulated as:
\begin{equation} \label{eq:col_term} 
\mathcal{L}_{\text{col.}} =  \sum_{i\in\mathcal{P}} \min _{j \in \mathcal{V}_{\text{f}}}\left\|\bV_{\text{h}}^{i} - \bV_{\text{f}}^{j}\right\|^{2}_{2} +\mathcal{L}_{\text{regDef}},
\end{equation}  
where $\mathcal{V}_{\text{f}}\,{\subset}\,\llbracket 1,M \rrbracket$ is the set of all the indices of the face vertices $\bV_{\text{f}}$. The term  $\mathcal{L}_{\text{regDef}}$ regularises the update of the deformation $\mathbf{p}$ from the perspective of edge lengths, neighbouring face angles and original deformation estimated by DefConNets. Let $l=\{l_{1},...,l_{x}\}$ and $\varphi=\{\varphi_{1},...,\varphi_{y}\}$ be vectors that consist of the edge lengths and the angles between the neighbouring faces of the face mesh, respectively. The formulation of $\mathcal{L}_{\text{regDef}}$ reads:
 \begin{equation} \label{eq:col_term_reg} 
\mathcal{L}_{\text{regDef}} \!= \!\sum_{i=1}^{x}s_{\text{edge}}^{i}\left\| l_i  - l_{0}\right\|^{2}_{2}  +\!\sum_{i=1}^{y}s_{\text{bend}}^{i}\left\| \varphi_i -  \varphi_{0}\right\|^{2}_{2}  +\! \left\| \mathbf{p} - \mathbf{p}_{0}\right\|^{2}_{2},
\end{equation} 
where $l_{0}$ and $\varphi_{0}$ denote the edge lengths and dihedral angles at rest and 
$\mathbf{p}_{0}$ is the displacements estimated by DefConNets in the previous step;  
$s_{\text{edge}}$ and $s_{\text{bend}}$ are, respectively, the edge and bending stiffness values that consider the underlying skull structure of a human head. 
The details of the stiffness computations are elaborated in Sec.~\ref{ssec:stiffness_on_head}. 

\begin{figure*}[t!] 
\includegraphics[width=1\linewidth]{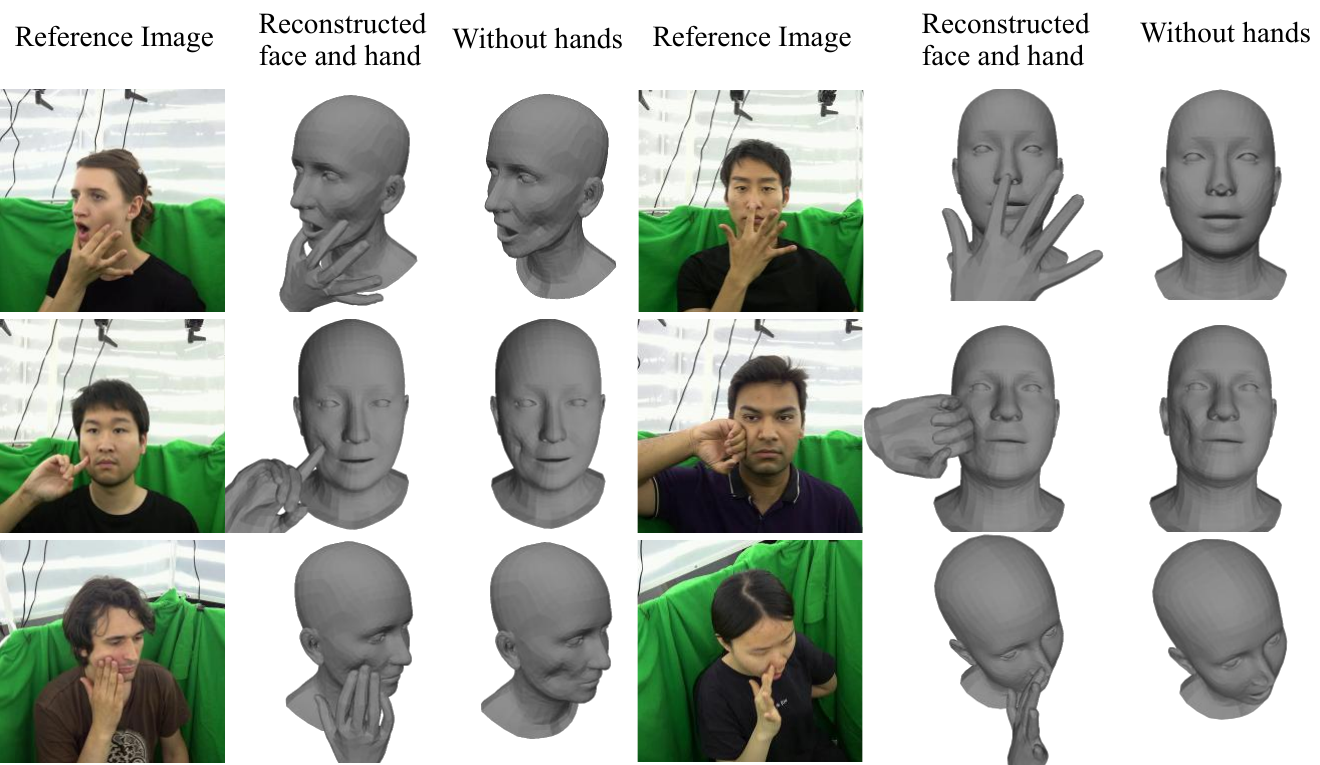} 
\caption{Example visualisations from our new hands+face 3D motion capture dataset with hand shape articulations non-rigid face deformation. 
The reconstructed 3D geometry shows plausible surface deformations thanks to the fitting optimisation combined with PBD. 
} \label{fig:dataset_showcase}
\end{figure*}

To further introduce the learned prior for the depth position of the hand, we train a conditional variational autoencoder (CVAE) \cite{sohn2015learning} -based depth prior network \textit{DePriNet} that is conditioned on the 2D key points. 
DePriNet is trained to reconstruct the 3D hand key points in a \textbf{canonical face frame}, as estimating the depth of hand and face in the camera frame only from monocular $2$D input is challenging due to the depth ambiguity (\eg $3$D hand and face with different combinations of focal lengths and depths can be projected onto the same position in the $2$D image). 
However, the hand positions relative to the face in the $3$D space are invariant to the depth in the camera frame; see Fig.~\ref{fig:depth_prior} for a schematic visualisation. 
We train DePriNet with the standard losses: 
\begin{equation} \label{eq:depth_term} 
\mathcal{L}_{\text{vae}} =   \left\|\bJ_{\text{h}}^{*}  - \hat{\bJ}_{\text{h}}^{*} \right\|^{2}_{2} + \operatorname{KL}\big(q(\mathbf{Z} \mid \hat{\bJ}_{\text{h}}^{*},\mathbf{\Theta}) \| \mathcal{N}(\mathbf{0},\mathbf{I})\big).
\end{equation}

The first term is a reconstruction loss to reproduce the ground-truth input hand joints in a canonical face frame $\hat{\bJ}_{\text{h}}^{*}\,{\in}\,\mathbb{R}^{K_{\text{h}} \times 3}$ and $\bJ_{\text{h}}^{*}\,{\in}\,\mathbb{R}^{K_{\text{h}} \times 3}$ denotes the output from the decoder network of DePriNet. 
The second loss term penalises the deviation of the latent vector $\mathbf{Z}\,{\in}\,\mathbb{R}^{50}$ distribution from a standard normal distribution $\mathcal{N}(\mathbf{0},\mathbf{I})$ using the Kullback-Leibler divergence loss $\operatorname{KL}(\cdot \| \cdot)$. Latent $\mathbf{Z}$ is sampled from a Gaussian distribution whose mean and variance are estimated from the encoder network $q(\cdot)$ of DePriNet. At test time, we use the decoder network $p(\cdot)$ of DePriNet to output depth candidates of the hand positions that are integrated into the depth prior loss $\mathcal{L}_{\text{depth}}$ in the global fitting optimisation:
\begin{gather} 
\label{eq:depth_prior} 
\mathcal{L}_{\text{depth}} =\sum_{i=1}^{u} w_{i}\left\|\bJ_{\text{h}}^{z} - \mathbf{T}(\bJ_{\text{h},i}^{*} )\right\|^{2}_{2}, \\
\text{where~}
    w_{i} = 1 - \frac{\eta_{i} - \min(\eta)}{\max(\eta) - \min(\eta)}, \;\; \eta_{i} =|\mathbf{Z}^{i}|_{1},
\end{gather}
$\bJ_{\text{h}}^{z}$ denotes the $z$-value of the hand 3D keypoints $\bJ_{\text{h}}$ that corresponds to the depth axis in the camera frame, and $\mathbf{T}(\cdot)$ is a transformation from the canonical face space to the camera frame that consists of the rotation and translation of the face model (that are also simultaneously obtained in this global fitting optimisation); 
$\bJ_{\text{h},i}^{*}$ is the $i$-th sample obtained from the decoder $p(\cdot)$ given $u =100$ latent vectors ${\sim}\mathcal{N}(\mathbf{0},\mathbf{I})$ and the conditioning vector $\mathbf{\Theta}$ that consists of face and hand $2$D keypoints with corresponding confidence values as well as the face $3$D rotation in the camera frame in $6$D representation \cite{zhou2019continuity}. 
Note that $2$D key points of the face and hands are translated to be a face-root relative representation for the conditioning. 
The conditioning 3D head rotation is obtained during the optimisation \eqref{eq:fitting_opt}. 
Each generated sample is weighted by the scalar $w$ that has the higher value 
the closer the corresponding latent vector $\mathbf{Z}$ is to zero (\ie a statistically more likely sample). 
We utilise the two independent DePriNets of the same architecture for the left and right hands. 
After minimising the objective that combines all these loss terms,
we obtain the final 3D head and hand reconstructions with plausible deformations and interactions. 
The significance of each loss term is evaluated in Sec.~\ref{sec:evaluations}. 
The final deformed face vertices $\bV^{*}_{\text{f}}$ are obtained by simply adding the updated deformations $\mathbf{p}$ to the face model parameterised by $\bPhi_{\text{f}}$, \ie $\bV^{*}_{\text{f}} = \mathcal{F}(\bPhi_{\text{f}}) + \mathbf{p}$.

\subsection{Architectures of Our Networks}\label{ssec:net_details} 

Our \method\ comprises several components (Fig.~\ref{fig:schematic_draw}). 
We employ \cite{bulat2017far} and \cite{lugaresi2019mediapipe} for 2D keypoint and bounding box estimation of the face and hand, respectively. 
The \textit{DefConNet} is composed of two encoders and three decoders. 
The encoders for the cropped face and hand images follow the ResNet-18 architecture \cite{resnet}. 
The decoders, sharing the same architecture, estimate per-vertex deformations and contact labels for the face and hand. 
Each of them includes three fully connected layers with leaky ReLU activation \cite{maas2013rectifier} and their hidden layer dimensions equal to $1024$. 
We duplicate \textit{DefConNet} for both hands and compute the union of the face deformations and contacts before the final global fitting optimisation. 
The \textit{DePriNet} is a variational autoencoder \cite{kingma2013auto}, consisting of three linear fully connected layers with batch normalisations, ReLU activations \cite{agarap2018deep}, a latent dimension of $50$ and hidden size of $128$ for both encoders and decoders. 

\section{Dataset}\label{sec:dataset}

In this work, we build a new markerless multi-view dataset for 3D hand-face interactions for method training and evaluation. 
It contains eight subjects---captured with 15 SONY DSC-RX0 cameras at $50$ fps (\textit{i.e.,} from $15$ different viewpoints)---along with the corresponding reference 3D geometries of a right hand and head, including surface deformations of the head represented as per-vertex displacements. 
In total, the dataset contains $100$K frames, see Table \ref{tab:dataset_breakdown} for the details. 
Each actor performs seven different actions with three different facial expressions.
For each captured view, the background masks are obtained using \cite{BMSengupta20}. The bounding boxes (for the hands and the faces) and $2$D key points (for the faces), are obtained using \cite{lugaresi2019mediapipe} and \cite{bulat2017far}, respectively. 
In the remainder of this section, 
we elaborate on our dataset generation pipeline; see Fig.~\ref{fig:data_pipeline} for the overview. 
The first step of the pipeline, \textit{i.e.,} multiview template fitting, is explained In Sec.~\ref{ssec:data_pipeline}. 
Next, to obtain a reasonable stiffness value that considers the underlying skull structure of a human face, we introduce a simple but effective skull-skin distance (SSD) approach in Sec.~\ref{ssec:stiffness_on_head}. 
The computed stiffness values are further utilised in the deformable object simulation relying on \textit{position based dynamics (PBD)}, and we obtain the final $3$D geometry with plausible interactions arising from hand-face interactions (Sec.\ref{ssec:pbd}). 

\subsection{Multiview Template Fitting}\label{ssec:data_pipeline} We first solve the $2$D keypoint reprojection-based fitting optimisation to obtain the MANO \cite{MANO:SIGGRAPHASIA:2017} and FLAME model \cite{FLAME:SiggraphAsia2017} parameters, so that the hand and face shapes match the multiview $2$D keypoints with known intrinsic and extrinsic calibrations. 
The objective for the face fitting encompasses \eqref{eq:reproj} and \eqref{eq:reg}. 
For the hand, we also minimise \eqref{eq:reproj} and \eqref{eq:reg} with the modification that \eqref{eq:reproj} is applied on the hand $3$D joints $\bJ_{\text{h}}$, and \eqref{eq:reg} is applied on the hand shape $\bbeta_{\text{h}}$, velocity and acceleration of hand vertices, excluding the expression loss term $\big\|\bPsi\big\|_2^{2}$. 
However, 
FLAME 
does not model the surface deformation caused by the interactions, which can result in physically implausible collisions; see the red circle in Fig.~\ref{fig:data_pipeline}-(1). 
We address this limitation by integrating into our tracking pipeline a deformable object simulator relying on position-based dynamics (PBD) \cite{pbd2007}. 
Our approach assumes non-homogeneous stiffness values of the human face, and we describe next how we obtain those.

\begin{table}[t] 
  \caption{Details of our new dataset. This dataset contains several types of data including pseudo ground truth of $3$D surface deformations represented as $3$D displacement vectors for seven different actions with three different facial expressions performed by eight subjects. The ``Age'' signifies the age range, whereas the number in the brackets means the corresponding number of subjects.}\label{tab:dataset_breakdown} 
  \center
  \begin{tabular}{cc}
    \toprule
    \textbf{Characteristic} & \textbf{Value/Description} \\  \midrule
    Number of subjects & 8 \\   
    Number of views & 16 \\  
    Total Number of Frames & 100 K \\  
    Ethnicity  & 5 Asian,  3 Caucasian \\   
     Gender  & 6 male,  2 female  \\ 
    Age  & 20 - 29 (5), 30 - 39 (3)  \\  
    Facial expressions & neutral, open mouth, smiling \\ \midrule
    Action types &  \makecell{ 
  poking a cheek (open hand) \\ 
  poking a cheek (pointing hand) \\
  punching a cheek\\
  pushing a cheek with a palm \\
  rubbing a cheek \\ 
  pinching a chin \\
  touching nose front \\
  touching nose from side  \\  
    } \\    \midrule 
    
Data types & \makecell{2D hand keypoints \\
2D face landmarks \\ 
RGB videos \\ 
foreground segmentation masks \\
hand-face bounding box\\
3D mesh for hand and face\\
3D surface deformations} \\    \bottomrule 
  \end{tabular}
\end{table}

\subsection{Stiffness on a Head Mesh}\label{ssec:stiffness_on_head} 
Deformable object simulators require known material stiffness.
The stiffness of human face tissues is non-uniform, due to the rich mimic musculature and the skull anatomy.  
Therefore, assuming uniform stiffness in the whole face and head would result in physically implausible artefacts when running the simulation; see Fig.~\ref{fig:stiffness_combined} for the examples. 
We obtain the non-uniform stiffness values based on a simple but effective \textit{skin-skull distance (SSD)} assumption. 
It is based on the assumption that our face and head region tend to have higher stiffness when the distance between the skin and skull surface is smaller (\eg forehead), and vice versa (\eg cheek). 
To compute SSD, we employ the mean skull and skin surface of a statistic model from \cite{achenbach2018multilinear}. 
The obtained tissue stiffness map is 
is upon our expectation and the corresponding pseudo-ground-truth deformations are used in quantitative experiments in Sec.~\ref{sec:evaluations}. 

\begin{figure}[t!] 
\includegraphics[width=1\linewidth] {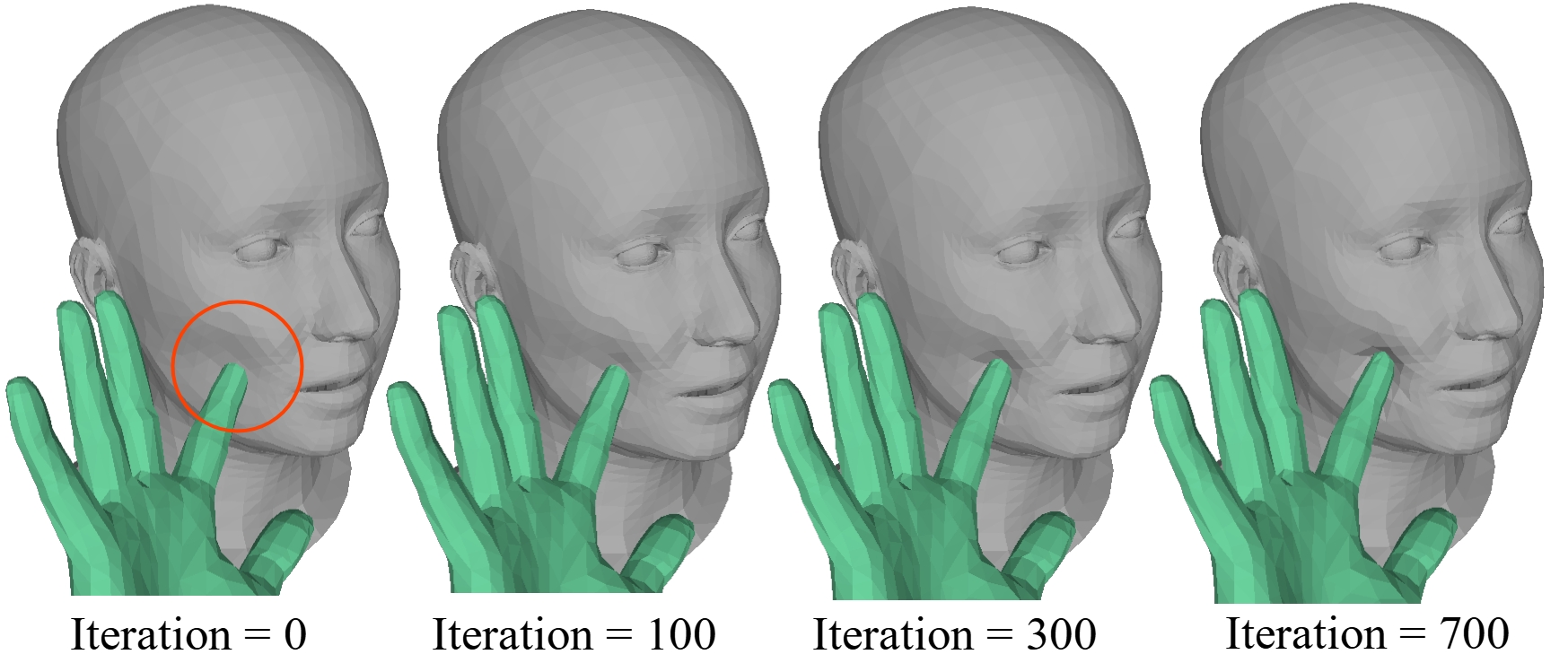} 
\caption{Visualisation of the effect of $\mathcal{L}_{\text{col.}}$ \eqref{eq:col_term}. Starting from the colliding hand and face poses (left-most visualisation), our non-rigid collision loss term effectively resolves the physically implausible inter-penetrations in the course of the optimisation. 
} 
\label{fig:non_rigid_optim} 
\end{figure} 
Let $\mathbf{D} = [d_{1}, ... ,d_{h} ]\,{\in}\,\mathbb{R}^{h}$ be a set of nearest distances between the skin and skull surfaces computed for all the $h$ skin vertices of \cite{achenbach2018multilinear}. The stiffness $s$ of the $i$-th skin vertex is calculated as follows: 
\begin{equation} 
    \mathbf{s}_{i} = ( 1 - \hat{d}_{i})^{b}, 
\end{equation}\label{eq:stiffness} where $\hat{d}$ is the normalised distance: 
\begin{equation} 
     \hat{d}_{i} =\frac{d_{i} - \min(\mathbf{D})}{\max(\mathbf{D}) - \min(\mathbf{D})}, 
\end{equation} 
with the operators $\min(\cdot)$ and $\max(\cdot)$ to compute the minimum and maximum values of the input vector; $b$ is empirically set to $4$. 
After computing the per-point stiffness $\mathbf{s}_{i}$, we transfer it to the FLAME head model by finding the corresponding vertices based on the nearest neighbour search after fitting the FLAME head model onto the skin surface model of \cite{achenbach2018multilinear}. 
In Fig.~\ref{fig:data_pipeline}-(2), we show the visualisation of the assigned stiffness values (more saturated green encodes lower stiffness).  
The assigned values are expected from the anatomical viewpoint (\eg high stiffness around the head region and low stiffness near the tip of the nose and cheeks). 
The edge and bending stiffness values in \eqref{eq:col_term_reg} are obtained by simply computing the average over the $s$ of vertices that form the edges and triangles.
\subsection{PBD-based Optimisation}\label{ssec:pbd} 
Position based dynamics (PBD) \cite{pbd2007} is a technique for simulating deformable objects, which gained popularity for its robustness and simplicity; 
it is widely used in game and physics engines. 
We utilise PBD to resolve implausible head-hand collisions which are challenging to address in a markerless motion capture setup due to constant occlusions at the interaction regions.
We utilise stretch constraint $C_{\text {stretch}}$, bending constraint $C_{\text {bend}}$ and collision constraint $C_{\text {collision}}$ in the PBD simulator. For each pair of connected vertices $\mathbf{p}_1$ and $\mathbf{p}_2$ in the mesh, $C_{\text {stretch}}$ is defined as follows: 
\begin{equation}
C_{\text{stretch}}\left(\mathbf{p}_1, \mathbf{p}_2\right)=\left|\mathbf{p}_1-\mathbf{p}_2\right|-l_0,
\end{equation} where  $l_0$ denotes the rest length of the edge between $\mathbf{p}_1$ and $\mathbf{p}_2$. For each pair of adjacent triangles ($\mathbf{p}_1, \mathbf{p}_3, \mathbf{p}_2$) and ($\mathbf{p}_1, \mathbf{p}_2, \mathbf{p}_4$), the definition of bending constraint $C_{\text {bend}}$ reads:
\begin{equation}
\begin{gathered}
C_{\text{bend}}\left(\mathbf{p}_1, \mathbf{p}_2, \mathbf{p}_3, \mathbf{p}_4\right)= \\
\operatorname{acos}\left(\frac{\left(\mathbf{p}_2-\mathbf{p}_1\right) \times\left(\mathbf{p}_3-\mathbf{p}_1\right)}{\left|\left(\mathbf{p}_2-\mathbf{p}_1\right) \times\left(\mathbf{p}_3-\mathbf{p}_1\right)\right|} \cdot \frac{\left(\mathbf{p}_2-\mathbf{p}_1\right) \times\left(\mathbf{p}_4-\mathbf{p}_1\right)}{\left|\left(\mathbf{p}_2-\mathbf{p}_1\right) \times\left(\mathbf{p}_4-\mathbf{p}_1\right)\right|}\right)-\varphi_0,
\end{gathered}
\end{equation} where $\varphi_0$ is the rest angle between the two triangles. Collision constraint $C_{\text {collision}}$ can be integrated for each vertex $\mathbf{p}$:
\begin{equation}
C_{\text {collision }}(\mathbf{p})=\mathbf{n}^T \mathbf{p}- h=0,
\end{equation} 
where $\mathbf{n}$ and $h$ are the normal of the colliding plane and the distance from the plane that $\mathbf{p}$ should maintain. After resolving collisions, we introduce friction as formulated in \cite{pbd2007} with $0.5$ for both kinetic and static friction coefficients. 

\begin{figure}[t!] 
\includegraphics[width=1\linewidth] {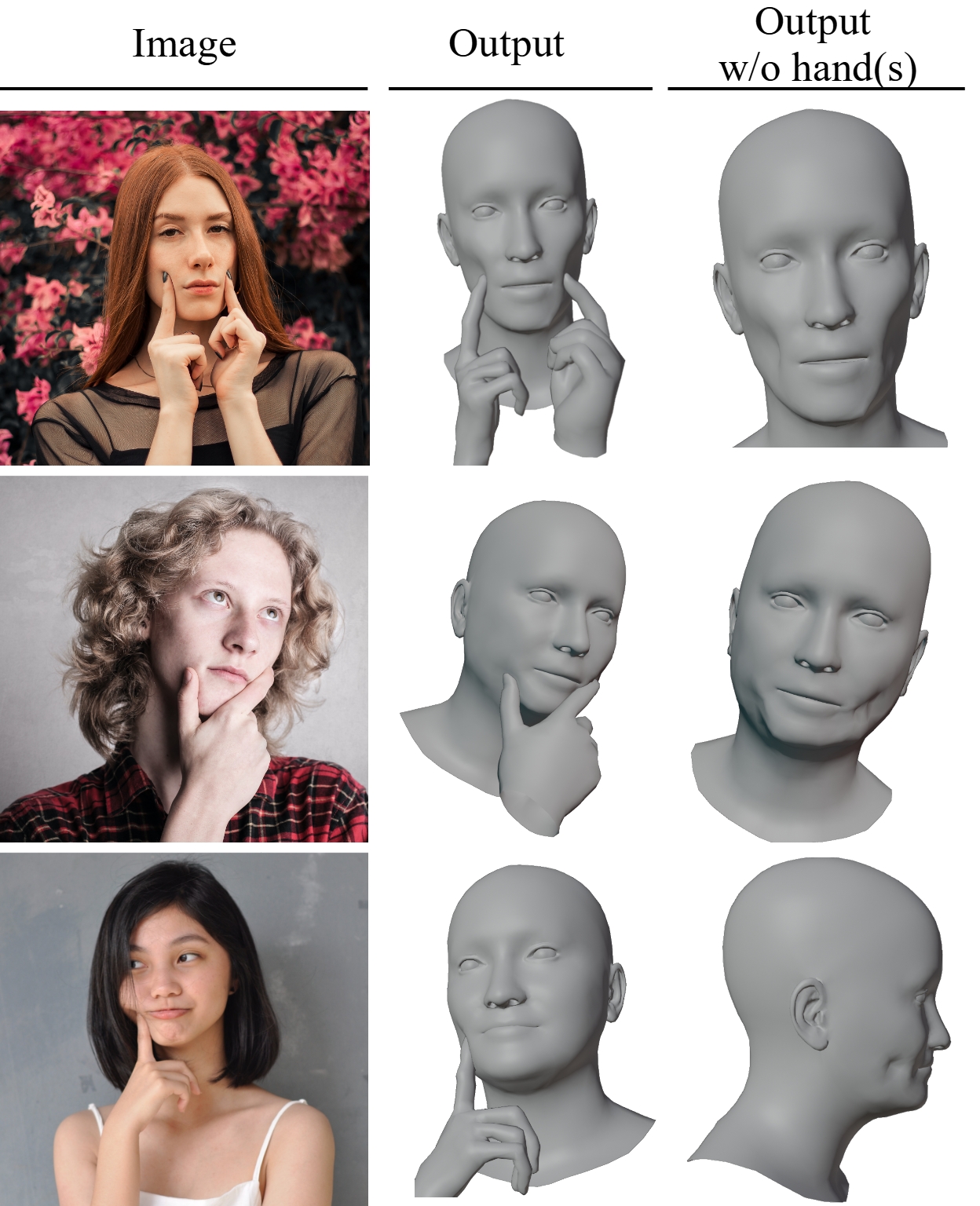} 
\caption{ 3D reconstructions on unseen identities in the wild. Our {\method} reasonably generalises across different identities and illuminations unseen during the training. The images are taken from \cite{Pexels}.
}  
\label{fig:wild} 
\end{figure} 

We also additionally introduce constraint $C_{\text {track}}$ for tracking the reference 3D motions obtained in Sec.~\ref{ssec:data_pipeline}. 
More specifically, this tracking constraint minimises the Euclidean distance between the vertex of the template mesh $\mathbf{p}$ and its corresponding vertex $\mathbf{p}_{\text{ref}}$ in the reference mesh from the previous multi-view fitting stage: 
\begin{equation}
C_{\text {stretch }}\left(\mathbf{p}, \mathbf{p}_{\text{ref}}\right)=\left|\mathbf{p}-\mathbf{p}_{\text{ref}}\right|.
\end{equation} 
For the simulation, we use the stiffness values obtained in Sec.\ref{ssec:stiffness_on_head}, and finally obtain the $3$D geometry of the interacting hand and face with the surface deformations (also see Fig.~\ref{fig:data_pipeline}-(3) for the example reconstruction).

\section{Evaluations}\label{sec:evaluations} 

 \begin{figure}[t!] 
\includegraphics[width=1\linewidth] {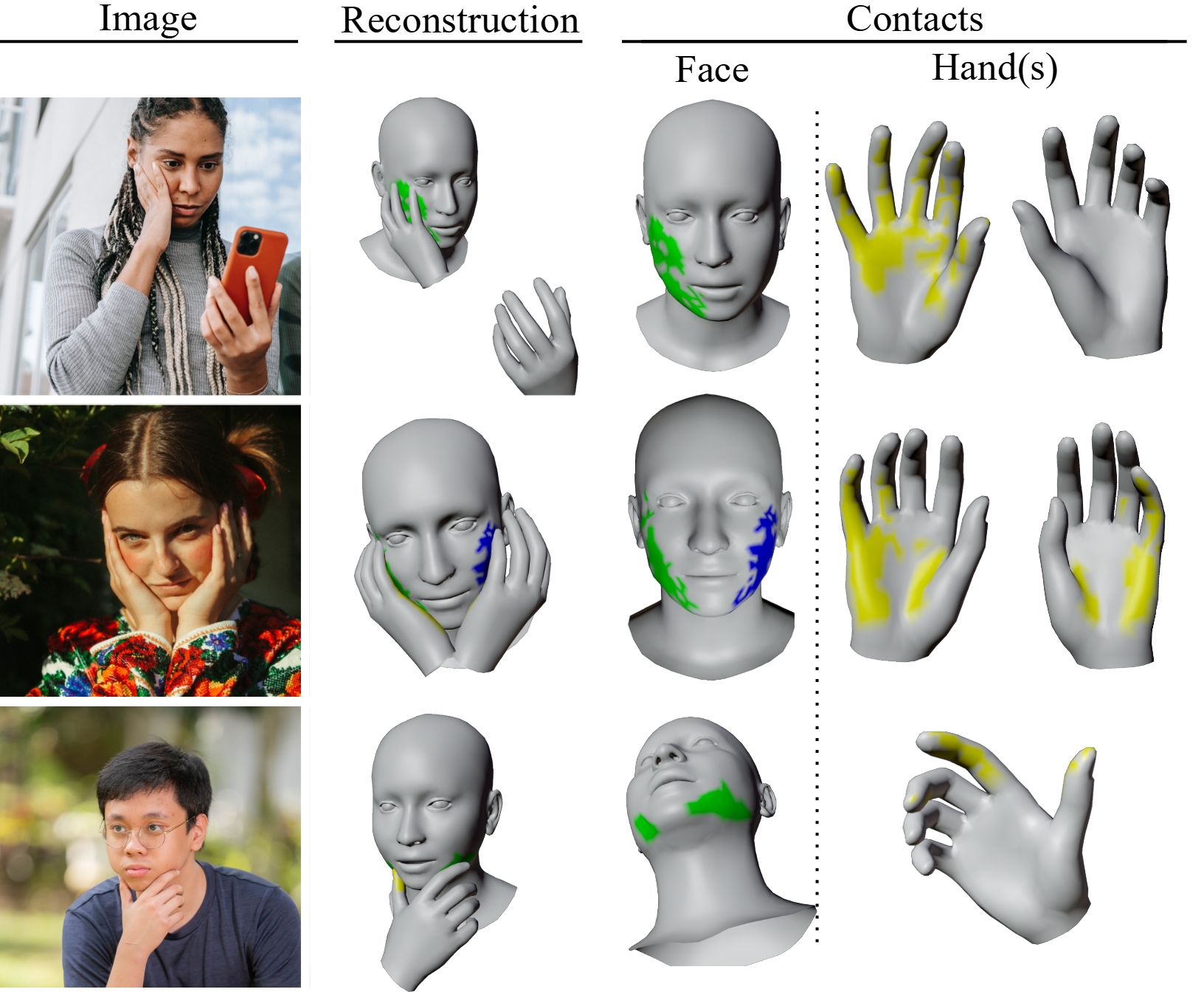} 
\caption{ Visualisations of the estimated contacts on in-the-wild images. The green and blue colours represent the face contacts regressed by the right- and left-hand DefConNet, respectively (see Fig.~\ref{fig:schematic_draw}). The yellow colour represents the contact regions on the hand(s). 
All estimations are reasonable. The images are taken from \cite{Pexels}.
} 
\label{fig:contacts}  
\end{figure}  
We next evaluate our \method\ on our new dataset. 
As there are no existing methods that address the same problem we tackle, we compare our method to a most closely related approach, \textit{i.e.,} a monocular full-body capture PIXIE \cite{PIXIE:3DV:2021} and its variants that reconstruct only hands and face independently, denoted as PIXIE (hand+face). 
We also compare to our benchmark method that includes hand-only \cite{lugaresi2019mediapipe} and face-only \cite{FLAME:SiggraphAsia2017} trackers.  

Note that in this method variant, DefConNet and non-rigid collision handling \eqref{eq:col_term} are deactivated. Our dataset contains separate training and testing sequences containing the same kinds of actions. We train our networks on the training sequences of $5$ different subjects and conduct the quantitative evaluations on $3$ different subjects unseen during the training. For the qualitative comparisons, we show the results of our data recording green studio and indoor sequences captured using a SONY DSC-RX0 camera. 
\subsection{Implementation and Training Details}\label{ssec:implementation} 
The neural networks were implemented in PyTorch \cite{NEURIPS2019_9015}. The evaluations and network training were conducted on a computer with an NVIDIA QUADRO RTX 8000 graphics card and AMD EPYC 7502P 32 Core Processor. The training was continued until convergence using Adam optimiser \cite{kingma2014adam} with a learning rate $3 \cdot 10^{-4}$.  \textit{DefConNet} models are trained until convergence which takes ${\approx}12$ hours. Since our dataset was captured with right-hand and face interactions,  
we flip the image and the corresponding $3$D ground-truth annotations and contact labels horizontally to obtain the input and ground truth for the left hand. 
For the global fitting optimisation, we set the loss term weights of \eqref{eq:fitting_head}, $\lambda_{\bbeta}= 1 \cdot 10^{-5}$, $\lambda_{\bPsi}=1 \cdot 10^{-3}$, $\lambda_{\dot{\bV}}= 3 \cdot 10^{-4}$, $\lambda_{\ddot{\bV}}= 3 \cdot 10^{-4}$. For \eqref{eq:fitting_hand}, we employed the following weights: $\lambda_{\text{touch}} = 0.1$,$\lambda_{\text{col.}} = 1.0$, $\lambda_{\text{depth}} = 3 \cdot 10^{-3}$, $\lambda_{\bbeta}= 1 \cdot 10^{-5}$ , $\lambda_{\dot{\bV}}= 3 \cdot 10^{-4}$, $\lambda_{\ddot{\bV}}= 3 \cdot 10^{-4}$. As the $2$D hand keypoint estimator \cite{lugaresi2019mediapipe} in our method estimates $3$D hand key points as well, we utilise them to initialise our hand pose by simply fitting the MANO hand model onto the 3D keypoints using inverse kinematics (Note that this step is optional.).

\begin{table*}[t] \caption{Comparisons of the $3$D reconstruction accuracy and plausibility of interactions. 
Lower PVE indicates higher 3D reconstruction accuracy. 
``$\dagger$'' denotes PVE after applying a translation on both the face and hand that translates the centre of the face mesh to the origin. 
Our \method\ shows the lowest error in PVE  and DefE metrics. 
In the plausibility measurements, lower Col. Dist. and higher Non. Col. indicate the lower magnitude of collisions and less frequent collisions, respectively (thus, more plausible interactions). 
Higher Touchness represents higher plausibility of the interaction that corresponds to the image input.
} 
\label{tab:threed_error_and_physical_plausibility}
\centering \scalebox{1.0}{ 
\begin{tabular}{c c c c  c c c  }\toprule   
&\multicolumn{2}{c}{3D Error}  & \multicolumn{4}{c}{Plausibility Measurement}    \\  
 \cmidrule(lr){2-3} \cmidrule(lr){4-7}
&   PVE [mm]$\downarrow$ &  PVE$\dagger$ [mm]$\downarrow$  &Col. Dist. [mm]$\downarrow$ &  Non. Col. [\%]$\uparrow$  &  Touchness [\%]$\uparrow$ &  F-Score [\%]$\uparrow$  \\  
\addlinespace[2pt] \midrule
 Ours       & \bf{11.9} &  \bf{9.65} &   1.03  & 83.6 & 96.6  & \bf{89.6} \\  
 Ours w/o $\mathcal{L}_{\text{touch}}$     & 17.4 &   15.2 &  6.83  &  68.7 & 78.5 &  73.2    \\ 
 Ours w/o $\mathcal{L}_{\text{col.}}$    & 15.7 &  12.9  & 14.4 & 59.6 & 87.7  & 71.0 \\  
 Ours w/o $\mathcal{L}_{\text{depth}}$    & 15.9  & 13.8  & 11.0 & 77.2 & 85.5  & 81.1 \\  
 Benchmark  & 18.9 & 17.7 & 19.3  & 64.2  &  73.2 &  68.4  \\
 PIXIE (hand+face)  &41.6 & 26.3 & 7.04 & 75.9  &  75.1 &  75.5 \\
 PIXIE & 51.9 &  39.7   & 0.11 & 97.1  &  51.8   & 67.6\\ \bottomrule 
 \end{tabular}}  
\end{table*}
\begin{table}[t!] \caption{$3$D deformation error comparisons. Lower DefE indicates higher 3D accuracy of the deformations. ``+'' indicates that DefE was computed only on deformations whose ground-truth deformation vector has a norm greater than $5$ [mm]. Our full method shows the lowest deformation error. Note that DefE and +DefE for related methods and benchmarks are computed using zero displacements as only our method outputs the per-vertex deformations (denoted with ``*'') .}\label{tab:deformations}
\centering \scalebox{1.0}{ 
\begin{tabular}{c  c c }\toprule   
&   DefE. [mm]$\downarrow$ &   +DefE. [mm]$\downarrow$   \\  
\addlinespace[2pt] \midrule
 Ours               &   \bf{0.08} & \bf{2.28}   \\ 
 Ours w/o refinement &  0.09  &   2.35  \\  
 Benchmark  &   0.13*& 7.28*\\
 PIXIE (hand+face)  &  0.13* & 7.28*\\ 
 PIXIE  &  0.13* & 7.28*  \\ \bottomrule 
\end{tabular}}
\end{table}

\subsection{Qualitative Evaluations}\label{ssec:qualitative} 
\begin{figure*}[t!] 
\includegraphics[width=0.88\linewidth] {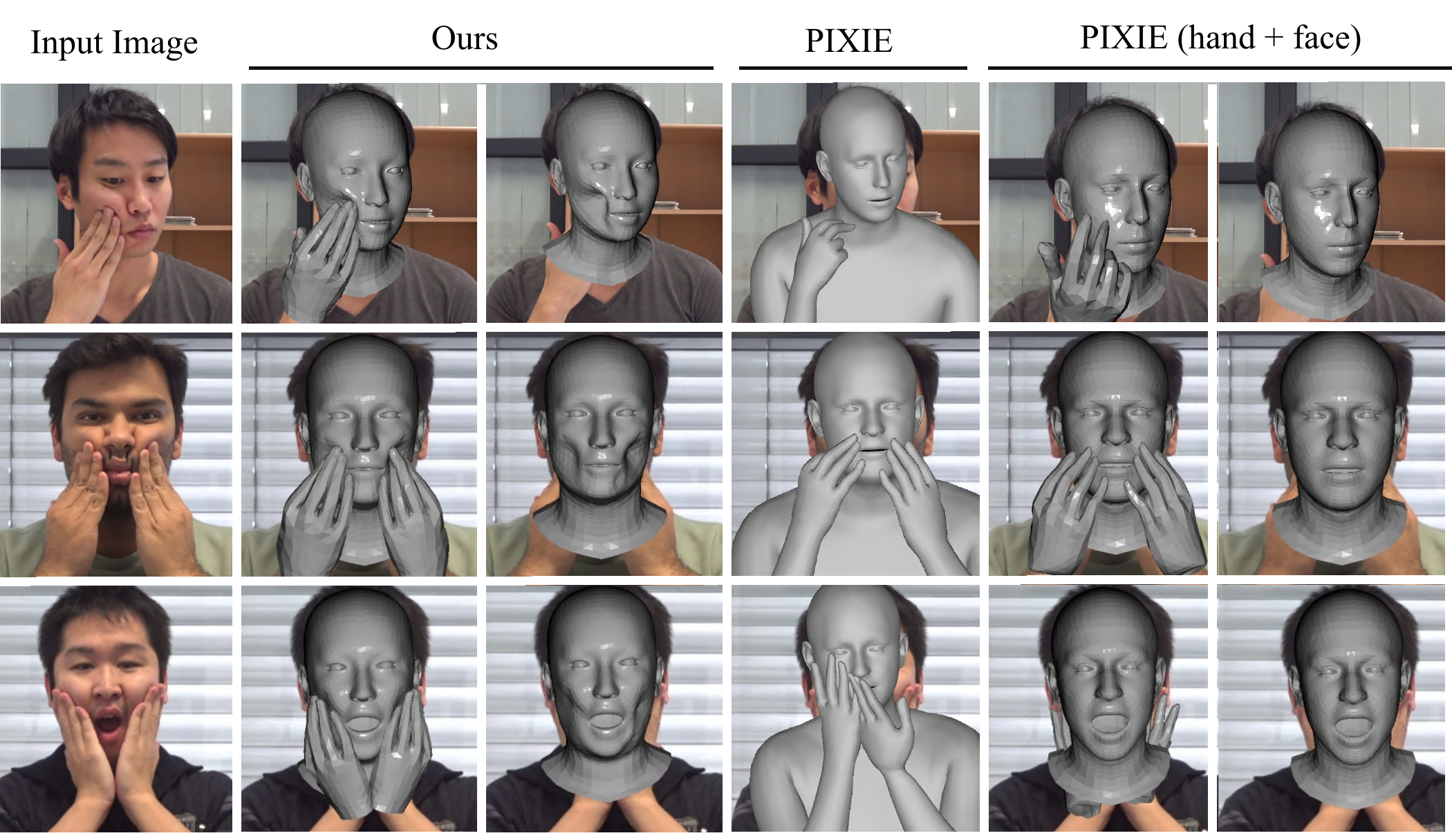} 
\caption{ Visualisations of the experimental results by our method, PIXIE \cite{PIXIE:3DV:2021} and hand-face-only mode of PIXIE for indoor scenes. Our results are plausible and represent expressive facial deformations, whereas the other works show inaccurate interactions and lack deformations. 
} 
\label{fig:mini_qualitative} 
\end{figure*}

Our supplementary video shows comparisons of our results with results of PIXIE (hand+face) \cite{PIXIE:3DV:2021} as well as the benchmark methods in a studio and an indoor scene, \ie monocular hand \cite{lugaresi2019mediapipe} and face \cite{FLAME:SiggraphAsia2017} trackers operating independently. 
Only our method reconstructs face deformations caused by the interactions while showing much more accurate $3$D localisations of the hands and face compared to other approaches; see Fig.~\ref{fig:big_qualitative} and Fig.~\ref{fig:mini_qualitative} for the visualisations. 
In Fig.~\ref{fig:non_rigid_optim}, we also show an example visualisation of the non-rigid collision loss \eqref{eq:col_term} starting from colliding hand and face positions. 
While the optimisation progresses, the physically implausible collisions are resolved by plausibly deforming the face surface. 
Our qualitative results confirm that \method\ produces significantly more plausible hand-face interactions and natural face deformations from a single RGB video compared with others. 

\begin{figure}[t!] 
\includegraphics[width=1\linewidth] {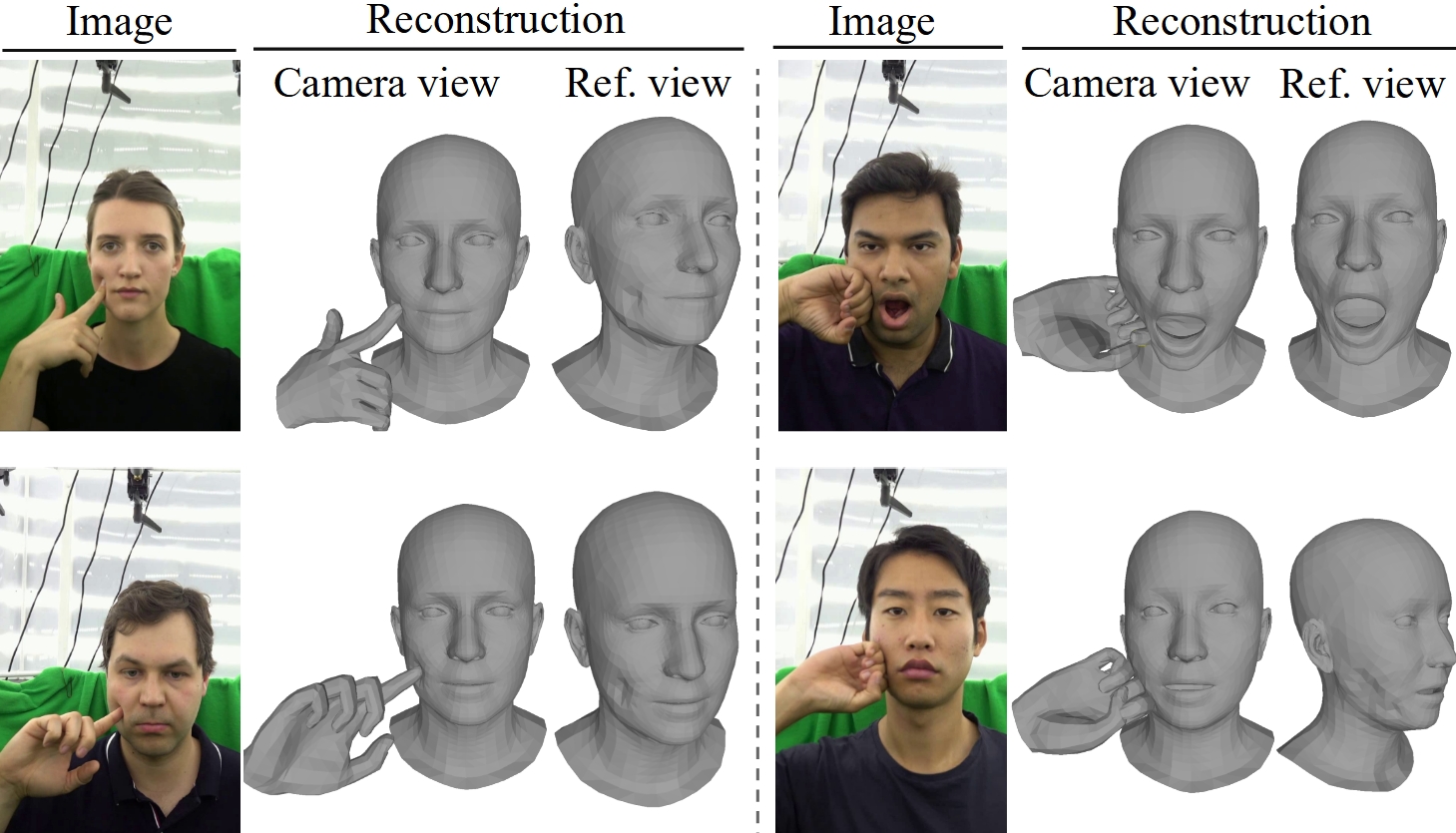} 
\caption{ 3D reconstructions on actions unseen during the training, \ie (left:) poking a cheek (pointing hand) and (right:) punching a cheek. 
} 
\label{fig:unseenposes} 
\end{figure} 
 To assess the generalisability of our {\method} across diverse identities and lighting conditions, we evaluate it on in-the-wild images;  see Fig.~\ref{fig:wild}. The reconstructed 3D shapes show plausible interactions with reasonable facial deformations. Furthermore, the estimated contacts showcased in Fig.~\ref{fig:contacts} %
faithfully mirror the contact regions evident in the input images. As a result, the final reconstructions show plausible hand-to-face interactions guided by the estimated contacts. 
 To further assess the generalisability of our method on unseen actions, we train our networks excluding  ``poking a cheek (pointing hand)'' and ``punching a cheek'' actions from the training dataset; the results for these actions are illustrated in  Fig.~\ref{fig:unseenposes}. Our method produces satisfactory results for ``poking a cheek (pointing hand)''. On the other hand, the exclusion of ``punching a cheek'' from the training dataset is a highly challenging scenario as no other actions in the training data contain interactions between the back side of the hand and the face. Given that our approach is neural and learning-based, 
such a substantial deviation from the training set can lead to inaccurate interactions in the results.

\begin{figure*}[t!] 
\includegraphics[width=1\linewidth] {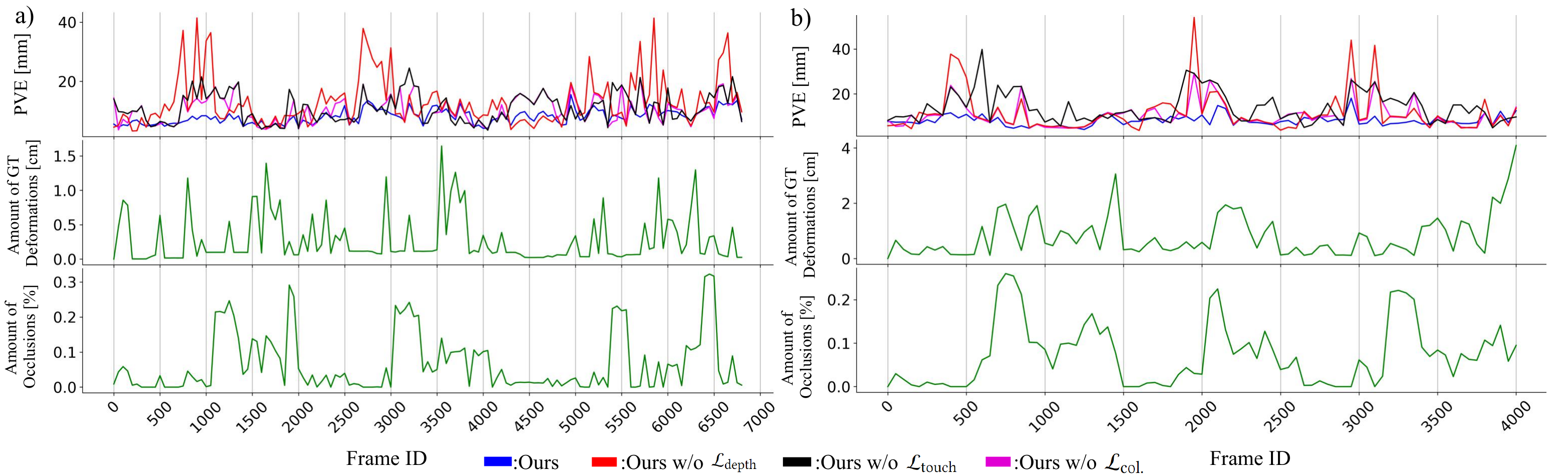} 
\caption{PVE plots for two exemplary test sequences (left: woman on top-left in Fig.~\ref{fig:dataset_showcase}; right: man on middle-right in Fig.~\ref{fig:dataset_showcase}) in relation to the degree of occlusions and deformations in the pseudo ground truth. 
Our full model is affected by the occlusions (the bottom row) substantially less than its ablated versions. 
} 
\label{fig:graphs} 
\end{figure*}

\subsection{Quantitative Evaluations}\label{ssec:quantitative}

To evaluate our algorithm from various perspectives numerically, we report multiple evaluation metrics. We calculate the 3D per vertex error (PVE) as an indicator of the 3D accuracy as well as the 3D deformation errors for our estimated face deformations. Additionally, we report the metrics of \textit{collision distance}, \textit{non-collision ratio} and \textit{touchness ratio} to quantify the physical plausibility of the reconstructed hands and faces. 
We also include the \textit{F-Score} to evaluate the overall plausibility of the reconstructions, taking into account both the occurrences of collisions and the correctness of the interactions. 
The specific details of each metric are elaborated as follows: 

\begin{itemize}[topsep=6pt,leftmargin=13pt]
  \item \textbf{Per vertex error (PVE)} measures the magnitude of the 3D error by computing the average Euclidean distances between the reconstruction and the ground-truth vertices. 
  We report the errors in the camera frame before and after applying a translation on the hand and face that aligns the centroid of the face with the origin of the coordinate frame, denoted as PVE and PVE$\dagger$, respectively. 
  Thus, PVE$\dagger$ measures the reconstruction quality focusing on the relative position of the hand w.r.t.~the head, which is important when judging the accuracy of the interactions. 
  \item \textbf{Deformation error (DefE)} measures the magnitude of the error by computing the average Euclidean distances between the estimated per-vertex 3D deformations and their pseudo ground truth. 
  We also report \textbf{+DefE} that computes DefE only for deformations with the corresponding ground-truth deformation vectors of norm greater than 5 [mm], \ie when non-negligible interactions are present. 
  Lower DefE~and +DefE indicate higher prediction accuracy of the deformations. 
  \item \textbf{Collision distance (Col. Dist.)} measures the collision distances averaged over the number of vertices and frames. A lower collision distance indicates a smaller magnitude of collisions throughout the sequence. 
  \item \textbf{Non-collision ratio (Non. Col.)} measures the ratio of the frames with no collisions between the hand and face over all sequence frames. 
  A higher non-collision ratio indicates fewer collisions in the reconstructed sequence. 
  \item \textbf{Touchness ratio} measures the ratio of frames over all the frames where contacts between face and hand are present in the prediction when there are face-hand contacts in our ground truth. 
  The hand vertices with the nearest distance from the face surface lower than 5 [mm] are considered in contact. 
  This metric exposes the presence of an artefact, namely the occurrence of face-hand interactions in the input frame while the hand does not make physical contact with the face in the reconstruction. 
  A higher ratio indicates more plausible reconstructions. 
  \item \textbf{F-Score} for Non. Col. and touchness ratio are also reported by computing the harmonic mean of the two (as these two metrics are complementary to each other). 
  It is very important to report F-Score, since each of these metrics in isolation is not meaningful 
  (\eg constant presence of hand-face collisions will result in perfect touchness ratio $100\%$; no presence of interaction throughout the sequence will make the perfect Non. Col. $100\%$). 
  A higher F-Score indicates a higher plausibility of the interactions in the reconstructions showing fewer occurrences of collisions and incorrect interactions. 
 
\end{itemize} 

\begin{figure*}[t!] 
\includegraphics[width=1\linewidth] {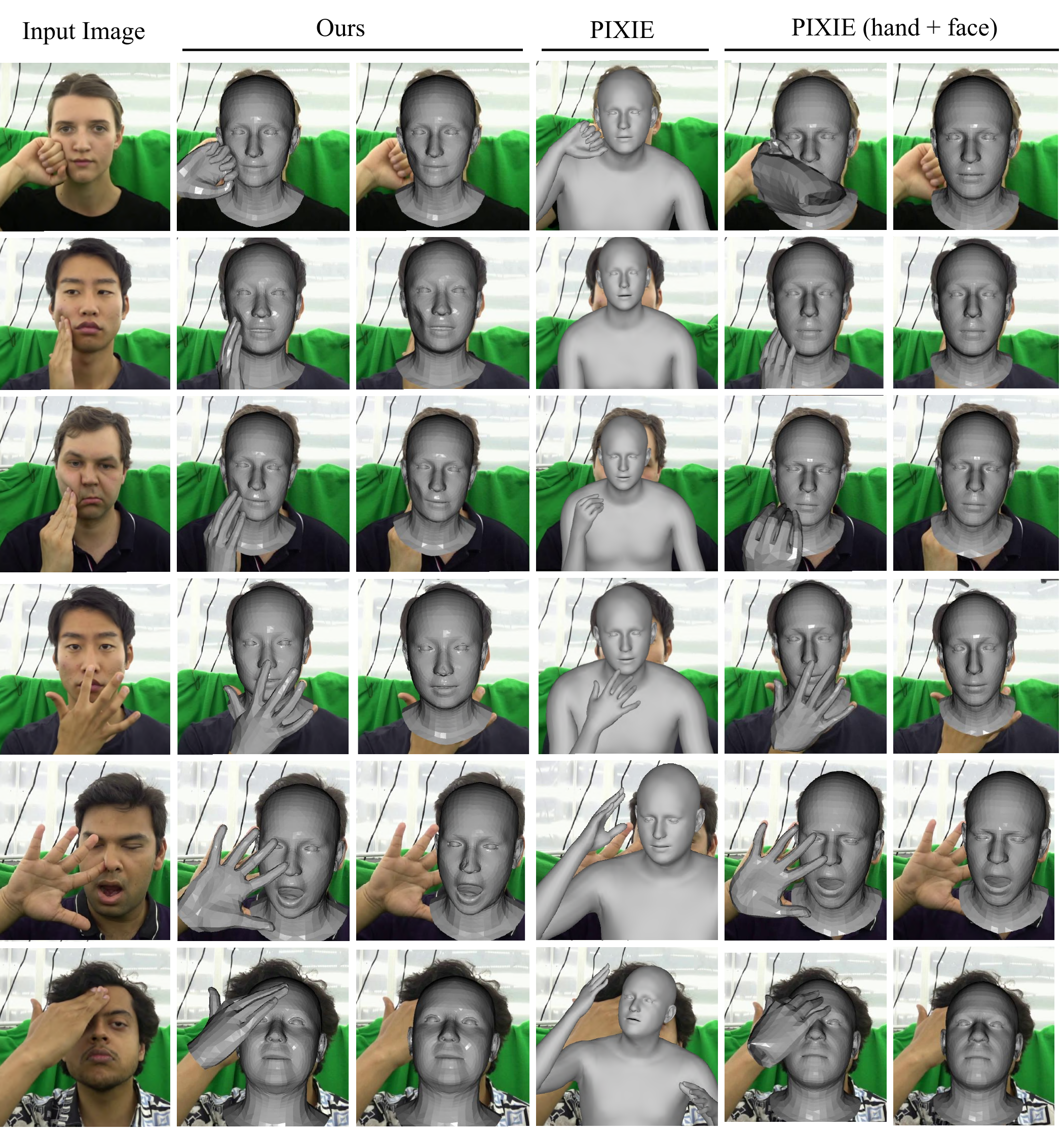} 
\caption{Visualisations of the experimental results by our method, PIXIE \cite{PIXIE:3DV:2021} and hand-face only mode of PIXIE. 
The PIXIE results (fourth column) frequently lack interactions between the hand and face, resulting in a low touchness ratio (Table \ref{tab:threed_error_and_physical_plausibility}). 
PIXIE (hand+face) in the fifth column shows collisions and lacks face-hand interactions as the method is agnostic to the latter.  
Our results (second column) exhibit natural interactions between the hand and face along with plausible face deformations 
(third column), which are not present in the results of the competing approaches 
(fourth and sixth columns). 
} 
\label{fig:big_qualitative} 
\end{figure*} 

\paragraph{3D Error Comparisons.} We report PVE in Table \ref{tab:threed_error_and_physical_plausibility}-(left) to evaluate the 3D accuracy of the reconstructed hand and face. 
Our \method\ shows the best performance scoring around $40\%$ less error compared with the second best method, benchmark (\cite{lugaresi2019mediapipe} + \cite{FLAME:SiggraphAsia2017}). 
We also report the 3D accuracies of the deformations; DefE and +DefE in Table \ref{tab:deformations}. 
To compute DefE for the related works, we simply provide zero deformations, as those methods do not model per-vertex deformations caused by interactions. 
For both DefE and +DefE, our method shows the lowest errors, \ie about $60\%$ lower errors for DefE and $40\%$ lower errors compared with others. 

\paragraph{Plausibility of Interactions.} In Table \ref{tab:threed_error_and_physical_plausibility}, we report Col. Dist., Non. Col., Touchness and F-Score. It is very important to show F-Score as Non. Col. and Touchness are \textit{complementary to each other}. 
Ours show low collision distances while showing quite high \textit{Touchness}, which indicates the highly plausible face-hand interactions that correspond to the input images, thus the best performance in F-Score. 
In contrast, PIXIE shows extremely low collision distances while showing much worse \textit{Touchness} compared with ours. 
This is because, in most cases, the reconstructed hand and face are wrongly not interacting with each other when they should be interacting; see Fig.~\ref{fig:big_qualitative} for the example reconstructions. 
The benchmark and PIXIE (hand+face) independently reconstruct the face and hands being agnostic of the interactions of those, therefore they show quite frequent collisions (high Col. Dist. and low Non. Col.) as well as incorrect interactions (Low Touchness), thus lower F-Score than ours. Given these metrics in Table \ref{tab:threed_error_and_physical_plausibility} and qualitative results in our video, \method\ shows the most plausible interactions in the reconstructed results compared with the related methods. 
\begin{table}[t] \caption{Perfomance measurement of our contact estimation component. Our method estimates reasonable contacts on face-hand surfaces only from RGB input, which are integrated into the final global fitting optimisation. The significance of the contacts is validated in Table \ref{tab:threed_error_and_physical_plausibility}. }\label{tab:contact}
\centering \scalebox{1.0}{ 
\begin{tabular}{c c c c  c }\toprule   
&  F-score $\uparrow$ &  Precision $\uparrow$ & Recall $\uparrow$ & Accuracy $\uparrow$   \\  
\addlinespace[2pt] \midrule
 face  & 0.57 & 0.69 & 0.49 & 0.99  \\  
 hand  & 0.47 & 0.62 & 0.39 & 0.98 \\ \bottomrule 
 \end{tabular}}  
\end{table}

\paragraph{Ablation Studies.} In Table \ref{tab:threed_error_and_physical_plausibility}, we show the ablation studies of the reconstructions denoted as ``Ours w/o  $\mathcal{L}_{\text{touch}}$'', ``Ours w/o  $\mathcal{L}_{\text{col.}}$ '' and ``Ours w/o  $\mathcal{L}_{\text{depth}}$ '' to assess the importance of each loss term. For both the $3$D accuracy and plausibility measurements, removing one loss term results in a severe performance decrease, which confirms all those loss terms contribute to higher $3$D localisations and improvement of interaction plausibilities. Additionally, in Table \ref{tab:deformations}, we also show the DefE and +DefE without updating the deformations in the final global fitting optimisation stage \ie direct output from the DefConNet denoted as ``Ours w/o refinement''. Our final global fitting optimisation improves the estimated deformations from DefConNet, reducing the DefE and +DefE by $10\%$ and $3\%$.

Fig.~\ref{fig:graphs} shows PVE plots for two test sequences from our dataset highlighting the stability of our results. 
\textit{Amount of occlusion} stands for the per-frame ratio of face pixels occluded by hand pixels from the camera view and \textit{amount of deformations} signifies the per-frame sums of deformations in the pseudo ground truth. 
We observe that the ablated versions of our method are starkly influenced by occlusions, which can be recognised with the help of peaks occurring at the frames with the (locally) largest deformations as well as the most significant occlusions. 
In contrast, our full model is affected by the occlusions substantially less and its curve has a smaller standard deviation of PVE, which verifies the importance of each loss term.

\paragraph{Contact Estimations.} To our knowledge, there are no existing works that estimate the contacts on hand-face surfaces from RGB inputs. Nonetheless, we report the performance of the contact estimation of our method for comparison on Table \ref{tab:contact}. Note that although estimating contact vertices only from RGB inputs is a highly challenging problem, our \method\ estimates reasonable contacts that significantly improve the $3$D localisation as validated in Table \ref{tab:threed_error_and_physical_plausibility}.

\section{Discussions and Limitations}
Our \method\ captures plausible 3D deformations along with hand-face interactions solely from a monocular RGB video, effectively reducing unnatural collisions and non-touching artefacts. While our method is the first to address this problem set, it does have certain limitations. Our network learns from a newly created dataset computed using Position-Based Dynamics (PBD) with a skull-skin-distance (SSD) approach combined with the multi-view markerless motion capture setup. PBD is widely utilised in modern physics engines, ensuring that our pseudo-ground truth deformations are plausible. However, it may introduce some discrepancies between the actual deformations and calculated deformations as this PBD-based approach does not integrate visual information such as photometric loss. Nevertheless, we believe this approach to be satisfactorily accurate to obtain plausible deformations although the visual information is not reliable at the interaction regions due to the constant occlusions, which is verified in our qualitative experiments. 
Our method employs PCA-based parametric face and hand models. Consequently, the 3D reconstructions of both body parts maintain consistent topology though, as a downside, miss high-frequency details such as wrinkles or blood vessels. 
Lastly, our method primarily focuses on handling pushing actions (\eg pushing or poking cheeks).  Furthermore, it is important to note that object-hand-face interactions, which fall outside the scope of our research, can be addressed in future studies.
\section{Conclusions}\label{sec:Conclusions} 
\method\ is the first monocular RGB-based approach for deformation-aware $3$D hand-face motion capture. 
Our method captures non-rigid face surface deformations arising from various hand-head interactions. 
It regards the human head anatomy (\textit{i.e.,} skull-skin distance used to calculate non-uniform facial tissue stiffness), detects hand-head contacts 
and is trained on a new dataset of facial performances. 
In the comprehensive experiments, \method\ demonstrates the highest $3$D 
reconstruction (in terms of PVE) and plausibility metrics (in terms of F-score) 
among all compared methods. 
Especially significant are the advancement in terms of PVE compared to the most closely related previous method (roughly fourfold error reduction) and qualitative improvements in the estimated 3D geometry, which opens up many possibilities for downstream applications (\textit{e.g.,} next-generation telepresence systems). 

\begin{acks}
We are very grateful to Weipeng Xu for his insightful discussions. All data captures were performed by MPII. The authors from MPII were supported by the ERC Consolidator Grant 4DRepLy (770784) and VIA research center. We acknowledge the support from Valeo.
\end{acks}
\bibliographystyle{ACM-Reference-Format}
\bibliography{sample-bibliography}
\end{document}